\pdfoutput=1

\documentclass[%
 reprint,
 amsmath,amssymb,
 aps,
]{revtex4-1}

\usepackage{natbib}
\usepackage{graphicx}
\usepackage{dcolumn}
\usepackage{bm}
\usepackage{hyperref}

\usepackage{capt-of}

\usepackage[page, title]{appendix}

\usepackage{setspace}
\usepackage{lineno}
\usepackage[section]{placeins}
\usepackage{afterpage}

\extrafloats{30}

\graphicspath{{figures/}}

\begin{document}

\preprint{APS/123-QED}


\title{Divergent discourse between protests and counter-protests:\\ \#BlackLivesMatter and \#AllLivesMatter}

\author{Ryan J. Gallagher}
\email{ryan.gallagher@uvm.edu}

\author{Andrew J. Reagan}

\author{Christopher M. Danforth}

\author{Peter Sheridan Dodds}

\affiliation{Department of Mathematics and Statistics, Computational Story Lab, \\ Vermont Complex Systems Center, and Vermont Advanced Computing Core \\ The University of Vermont, Burlington, VT 05405}

\date{\today}


\begin{abstract} 
Since the shooting of Black teenager Michael Brown by White police officer Darren Wilson in Ferguson, Missouri, the protest hashtag \#BlackLivesMatter has amplified critiques of extrajudicial killings of Black Americans. In response to \#BlackLivesMatter, other Twitter users have adopted \#AllLivesMatter, a counter-protest hashtag whose content argues that equal attention should be given to all lives regardless of race. Through a multi-level analysis of over 860,000 tweets, we study how these protests and counter-protests diverge by quantifying aspects of their discourse. We find that \#AllLivesMatter facilitates opposition between \#BlackLivesMatter and hashtags such as \#PoliceLivesMatter and \#BlueLivesMatter in such a way that historically echoes the tension between Black protesters and law enforcement. In addition, we show that a significant portion of \#AllLivesMatter use stems from hijacking by \#BlackLivesMatter advocates. Beyond simply injecting \#AllLivesMatter with \#BlackLivesMatter content, these hijackers use the hashtag to directly confront the counter-protest notion of ``All lives matter.'' Our findings suggest that Black Lives Matter movement was able to grow, exhibit diverse conversations, and avoid derailment on social media by making discussion of counter-protest opinions a central topic of \#AllLivesMatter, rather than the movement itself.
\end{abstract}

\maketitle

\section{Introduction}

Protest movements have a long history of forcing difficult conversations in order to enact social change, and the increasing prominence of social media has allowed these conversations to be shaped in new and complex ways. Indeed, significant attention has been given to how to quantify the dynamics of such social movements. Recent work studying social movements and how they evolve with respect to their causes has focused on Occupy Wall Street \cite{conover2013digital, caren2011occupy, deluca2012occupy}, the Arab Spring \cite{howard2011opening}, and large-scale protests in Egypt and Spain \cite{papacharissi2012affective, anduiza2014mobilization}. The network structures of movements have also been leveraged to answer questions about how protest networks facilitate information diffusion \cite{gonzalez2016networked}, align with geospatial networks \cite{conover2013geospatial}, and impact offline activism \cite{gonzalez2011dynamics, steinert2015online, larson2016social}. Both offline and online activists have been shown to be crucial to the formation of protest networks \cite{christensen2011political, barbera2015critical} and play a critical role in the eventual tipping point of social movements \cite{Borge-Holthoefere1501158, qi2016open}.

The protest hashtag \#BlackLivesMatter has come to represent a major social movement. The hashtag was started by three women, Alicia Garza, Patrisse Cullors, and Opal Tometi, following the death of Trayvon Martin, a Black teenager who was shot and killed by neighborhood watchman George Zimmerman in February 2012 \cite{garza2014herstory}. The hashtag was a ``call to action'' to address anti-Black racism, but it was not until November 2014 when White Ferguson police officer Darren Wilson was not indicted for the shooting of Michael Brown that \#BlackLivesMatter saw widespread use. Since then, the hashtag has been used in combination with other hashtags, such as \#EricGarner, \#FreddieGray, and \#SandraBland, to highlight the extrajudicial deaths of other Black Americans. \#BlackLivesMatter has organized the conversation surrounding the broader Black Lives Matter movement and activist organization of the same name. Some have likened Black Lives Matter to the New Civil Rights movement \cite{harris2015next, kang2015our}, though the founders reject the comparison and self-describe Black Lives Matter as a ``human rights'' movement \cite{tometi2015black}.

Researchers have only just begun to study the emergence and structure of \#BlackLivesMatter and its associated movement. To date and to the best of our knowledge, Freelon et al. have provided the most comprehensive data-driven study of Black Lives Matter \cite{freelon2016beyond}. Their research characterizes the movement through multiple frames and analyzes how Black Lives Matter has evolved as a movement both online and offline. Other researchers have given particular attention to the beginnings of the movement and its relation to the events of Ferguson, Missouri. Jackson and Welles have shown that the initial uptake of \#Ferguson, a hashtag that precluded widespread use of \#BlackLivesMatter, was due to the early efforts of ``citizen journalists'' \cite{jackson2015ferguson}, and Bonilla and Rosa argue that these citizens framed the story of Michael Brown in such a way that facilitated its eventual spreading \cite{bonilla2015ferguson}. Other related work has attempted to characterize the demographics of \#BlackLivesMatter users \cite{olteanu2015characterizing}, how \#BlackLivesMatter activists affect systemic political change \cite{freelon2016quantifying}, and how the movement self-documents itself through Wikipedia \cite{twyman2016black}.

\begin{figure*}[!htb]
\includegraphics[scale=.39]{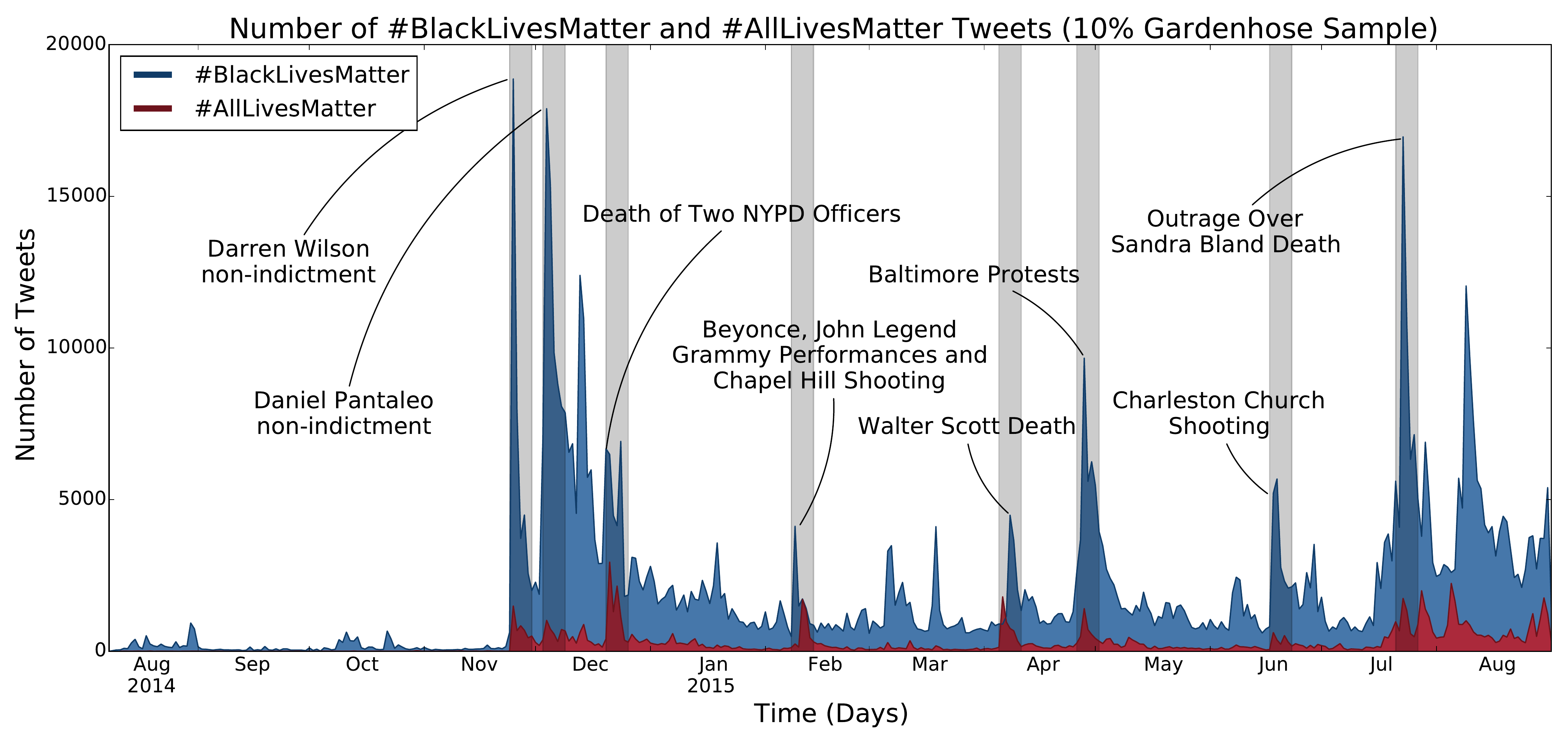}
\caption{Time series showing the number of \#BlackLivesMatter and \#AllLivesMatter tweets from Twitter's 10\% Gardenhose sample. The plot is annotated with several major events pertaining to the hashtags. Shaded regions indicate one-week periods where use of both \#BlackLivesMatter and \#AllLivesMatter peaked in frequency. These are the periods we focus on in the present study.}
\label{freq-time-series}
\end{figure*}

\#BlackLivesMatter has found itself contended by a relatively vocal counter-protest hashtag: \#AllLivesMatter. Advocates of \#AllLivesMatter affirm that equal attention should be given to all lives, while \#BlackLivesMatter supporters contend that such a sentiment derails the Black Lives Matter movement. The counter-hashtag \#AllLivesMatter has received less attention in terms of research, largely being studied from a theoretical angle. The phrase ``All Lives Matter'' reflects a ``race-neutral'' or ``color-blind'' approach to racial issues \cite{orbe2015black}. While this sentiment may be ``laudable,'' it is argued that race-neutral attitudes can mask power inequalities that result from racial biases \cite{husband2012don}. From this perspective, those who adopt \#AllLivesMatter evade the importance of race in the discussion of Black deaths in police-involved shootings \cite{rickford2016black, speight2015black}. To our knowledge, our work is the first to engage in a data-driven approach to understanding \#AllLivesMatter. This approach not only allows us to substantiate several broad claims about the use of \#AllLivesMatter, but to also highlight trends in \#AllLivesMatter that are absent from the theoretical discussion of the hashtag.

Given the popular framing of Black Lives Matter as the New Civil Rights movement, the movement and its oppositions are in and of themselves of interest to study. Furthermore, while qualitative study of \#AllLivesMatter has provided a theoretical framing of the hashtag, it is still unclear to what extent, if any, the All Lives Matter movement hijacked the conversations of \#BlackLivesMatter. Through the computational analysis of over 860,000 tweets, in this paper we comprehensively quantify the ways in which the protest and counter-protest discourses of \#BlackLivesMatter and \#AllLivesMatter diverge most significantly. Unlike previous studies of political polarization that focus on hashtag trends or curated lists of terms \cite{conover2012partisan, bode2015candidate, borge2015content, weber2013secular}, the methods we leverage take advantage of the entirety of textual data, thus giving weight to all protest sentiments that were voiced through these hashtags.

Through application of these methods at the word level and topic level, we first show that \#AllLivesMatter diverges from \#BlackLivesMatter through support of pro-law-enforcement sentiments. This places \#BlackLivesMatter and hashtags such as \#PoliceLivesMatter and \#BlueLivesMatter in opposition, a framing that is in line with the theoretical understanding of \#AllLivesMatter and that mimics how relationships between black protesters and law enforcement have been historically depicted. We then demonstrate that \#AllLivesMatter experiences significant hijacking from \#BlackLivesMatter, while  \#BlackLivesMatter exhibits rich and informationally diverse conversations, of which hijacking is a much smaller portion. These findings, which augment the theoretical discussion of the impact of the All Lives Matter movement, suggest that the Black Lives Matter movement was able to avoid being derailed by counter-protest opinions on social media by relegating discussion of such opinions to the counter-protest, rather than the movement itself.


\section{Materials and Methods}


\subsection{Data Collection}

We collected tweets containing \#BlackLivesMatter and \#AllLivesMatter (case-insensitive) from the period August 8th, 2014 to August 31st, 2015 from the Twitter Gardenhose feed. The Gardenhose represents a 10\% random sample of all public tweets. Our subsample resulted in 767,139 \#BlackLivesMatter tweets from 375,620 unique users and 101,498 \#AllLivesMatter tweets from 79,753 unique users. Of these tweets, 23,633 of them contained both hashtags. When performing our analyses, these tweets appear in each corpus.

Previous work has emphasized the importance of viewing protest movements through small time scales \cite{freelon2016beyond, jackson2015ferguson}. In addition, we do not attempt to characterize all of the narratives that exist within \#BlackLivesMatter and \#AllLivesMatter. Therefore, we choose to restrict our analysis to eight one-week periods where there were simultaneous spikes in \#BlackLivesMatter and \#AllLivesMatter. These one-week periods are labeled on Figure~\ref{freq-time-series} and are as follows:

\begin{enumerate}
\item \emph{November 24th, 2014}: the non-indictment of Darren Wilson in the death of Michael Brown. \cite{mcclam2014ferguson}
\item \emph{December 3rd, 2014}: the non-indictment of Daniel Pantaleo in the death of Eric Garner \cite{sanchez2014protests}.
\item \emph{December 20th, 2014}: the deaths of New York City police officers Wenjian Liu and Rafael Ramos \cite{mueller2014nypd}.
\item \emph{February 8th, 2015}: the Chapel Hill shooting and the 2015 Grammy performances by Beyonce and John Legend \cite{ahmed2015students, ramirez2015grammys}.
\item \emph{April 4th, 2015}: the death of Walter Scott \cite{elmore2015man}.
\item \emph{April 26th, 2015}: the social media peak of protests in Baltimore over the death of Freddie Gray \cite{graham2015mysterious}.
\item \emph{June 17th, 2015}: the Charleston Church shooting \cite{berenson2015everything}.
\item \emph{July 21st, 2015}: outrage over the death of Sandra Bland \cite{schuppe2015death}.
\end{enumerate}


\subsection{Entropy and Diversity}
\label{entropy-and-diversity}

A large part of our textual analysis relies on tools from information theory, so we describe these methods here and frame them in the context of the corpus. Given a text with $n$ unique words where the $i$th word appears with probability $p_i$, the Shannon entropy $H$ encodes ``unpredictability'' as
\begin{equation}
H 
	= - \sum_{i = 1}^n p_i \log_2 p_i.
\end{equation}
Shannon's entropy describes the unpredictability of a body of text, and so, intuitively, we say that a text with higher Shannon's entropy is less predictable than a text with lower Shannon's entropy. It can then be useful to think of Shannon's entropy as a measure of diversity, where high entropy (unpredictability) implies high diversity. In this case, we refer to Shannon's entropy as the Shannon index. We employ this raw diversity measure in Section \ref{word divergence} to examine the diversity of language surrounding individual words.

Of the diversity indices, only the Shannon index gives equal weight to both common and rare words \cite{jost2006entropy}. In addition, even for a fixed diversity index, care must be taken in comparing diversity measures to one another \cite{jost2006entropy}. In order to make linear comparisons of diversity between texts, we convert the Shannon index to an effective diversity. The effective diversity $D$ of a text $T$ with respect to the Shannon index is given by
\begin{equation}
D
	= 2^H
	= 2^{ \bigl(-\sum_{i = 1}^n p_i \log_2 p_i \bigr)}.
\label{effective-diversity}
\end{equation}
The expression in Eq.~\ref{effective-diversity} is also known as the perplexity of the text. The effective diversity allows us to accurately make statements about the ratio of diversity between two texts. For example, unlike the raw Shannon index, the effective diversity doubles in the situation of comparing texts with $n$ and $2n$ equally-likely words. We apply the effective diversity in Section \ref{conversational diversity}.


\subsection{Jensen-Shannon Divergence}

The Kullback-Leibler divergence is a statistic that assesses the distributional differences between two texts. Given two texts $P$ and $Q$ with a total of $n$ unique words, the Kullback-Leibler divergence between $P$ and $Q$ is defined as
\begin{equation}
D_{KL}(P || Q) = \sum_{i = 1}^n p_i \log_2 \frac{p_i}{q_i},
\end{equation}
where $p_i$ and $q_i$ are the probabilities of seeing word $i$ in $P$ and $Q$ respectively. However, if there is a single word that appears in one text but not the other, this divergence will be infinitely large. Because such a situation is not unlikely in the context of Twitter, we instead leverage the Jensen-Shannon divergence (JSD) \cite{lin1991divergence}, a smoothed version of the Kullback-Leibler divergence:
\begin{equation}
D_{JS}(P || Q) = \pi_1 D_{KL}(P || M) + \pi_2 D_{KL}(Q || M).
\end{equation}
Here, $M$ is the mixed distribution $M = \pi_1 P + \pi_2 Q$ where $\pi_1$ and $\pi_2$ are weights proportional to the sizes of $P$ and $Q$ such that $\pi_1 + \pi_2 = 1$. The Jensen-Shannon divergence has been previously used in textual analyses that range from the study of language evolution \cite{pechenick2015characterizing, pechenick2015language} to the clustering of millions of documents \cite{boyack2011clustering}.

The JSD has the useful property of being bounded between 0 and 1. The JSD is 0 when the texts have exactly the same word distribution, and is 1 when neither text has a single word in common. Furthermore, by the linearity of the JSD we can extract the contribution of an individual word to the overall divergence. The contribution of word $i$ to the JSD is given by
\begin{equation}
D_{JS,i}(P || Q) = 
	-m_i \log_2 m_i + \pi_1 p_i \log_2 p_i + \pi_2 q_i \log_2 q_i,
\end{equation}
where $m_i$ is the probability of seeing word $i$ in $M$. The contribution from word $i$ is 0 if and only if $p_i = q_i$. Therefore, if the contribution is nonzero, we can label the contribution to the divergence from word $i$ as coming from text $P$ or $Q$ by determining which of $p_i$ or $q_i$ is larger.

\section{Results}

\subsection{Word-Level Divergence}
\label{word divergence}

For each of the eight time periods, the collections of \#BlackLivesMatter and \#AllLivesMatter tweets are each represented as bags of words where user handles, links, punctuation, stop words, the retweet indicator ``RT,'' and the two hashtags themselves are removed. We then calculate the Jensen-Shannon divergence between these two groups of text, and rank words by percent contribution to the total divergence. We present the results of applying the JSD to the weeks following the non-indictment of Darren Wilson and the Baltimore protests in Figures~\ref{2014-11-24-wordshift}--\ref{2015-04-26-wordshift} and the remaining periods in Appendix Figures~\ref{2014-12-03-wordshift}--\ref{2015-07-21-wordshift}.

\begin{figure}[p!]
\centering
\includegraphics[scale = .215]{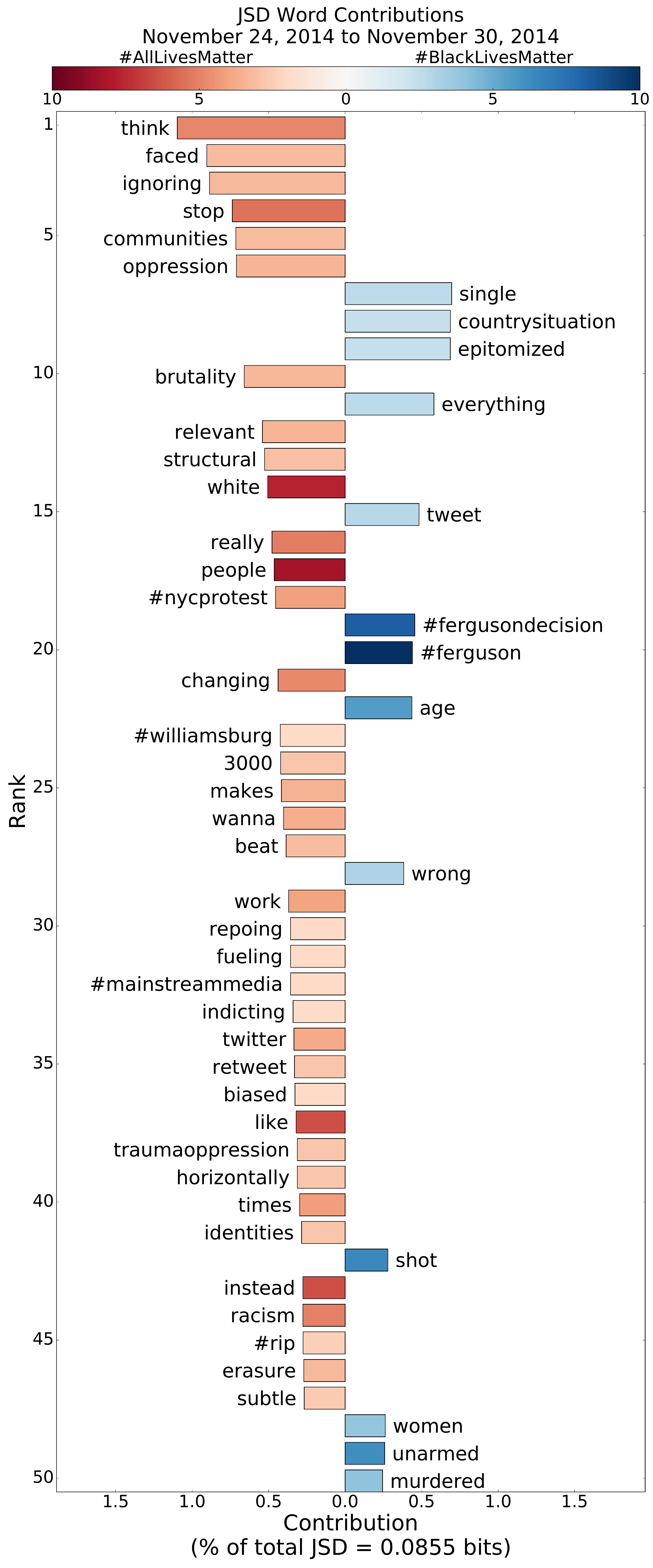}
\caption{Jensen-Shannon divergence word shift graph for the week following the non-indictment of Darren Wilson. All word contributions are positive percentages of the total divergence, where bars to the left and right indicate the word was more common in \#AllLivesMatter and \#BlackLivesMatter respectively. Shading indicate the Shannon index (in bits) of tweets containing the given word. Lighter shading indicates the contribution is due to one or several popular retweets. Darker shading indicates the contribution is due to the word being used throughout many different tweets.}
\label{2014-11-24-wordshift}
\end{figure}

\begin{figure*}
\centering
\begin{minipage}[t]{0.45\linewidth}
\includegraphics[scale = .215]{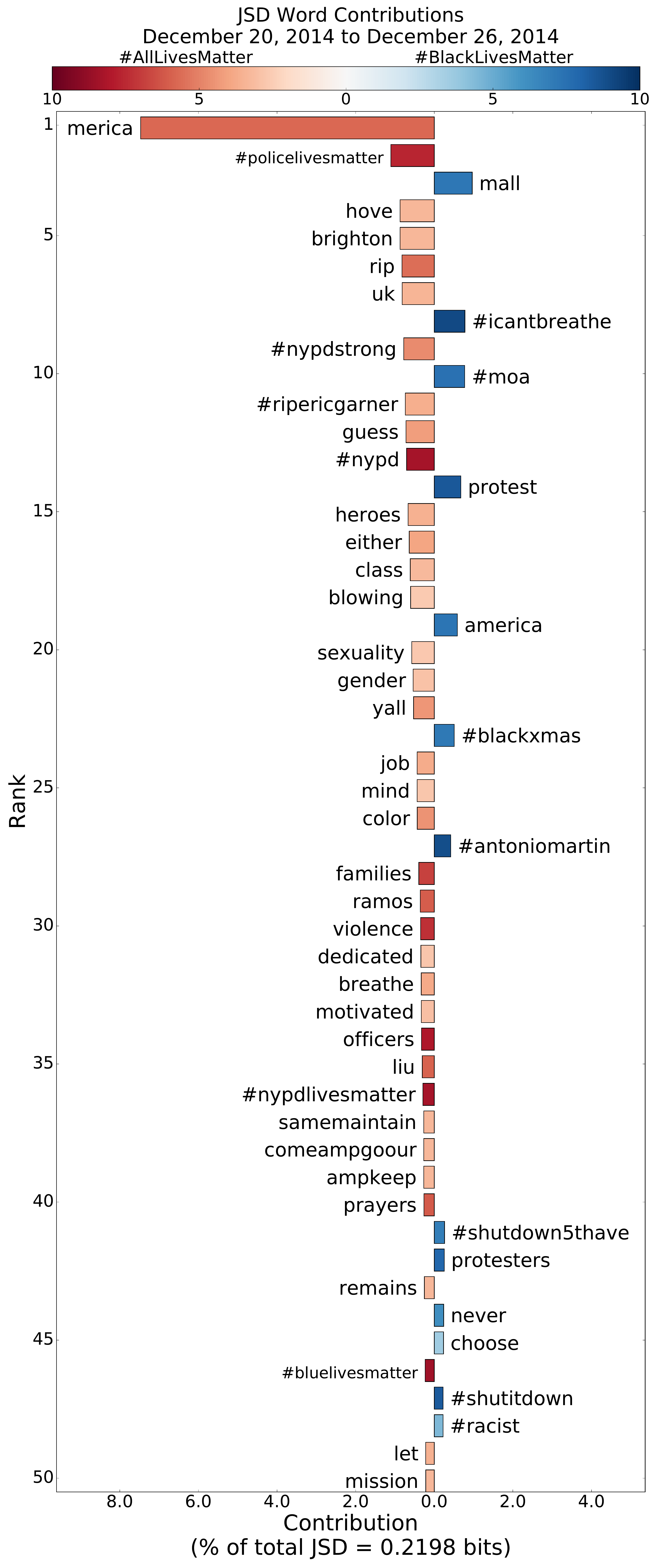}
\caption{Jensen-Shannon divergence word shift graph for the week following the death of two NYPD police officers. See main text and the caption of  Fig.~\ref{2014-11-24-wordshift} for an explanation of the word shift graphs. The diversity of ``merica'' appears to be high, however it is almost exclusively used in one popular retweet. This high diversity is due to alterations of the original retweet that added comments. Since these less popular, modified retweets contain different language from one another, \emph{and} the original tweet has no other words used in it, the diversity surrounding ``merica'' is artificially elevated. So, although uncommon, the diversity measure may be difficult to interpret in the case of popular one-word retweets.}
\label{2014-12-20-wordshift}
\end{minipage}
\quad
\begin{minipage}[t]{0.45\linewidth}
\includegraphics[scale = .215]{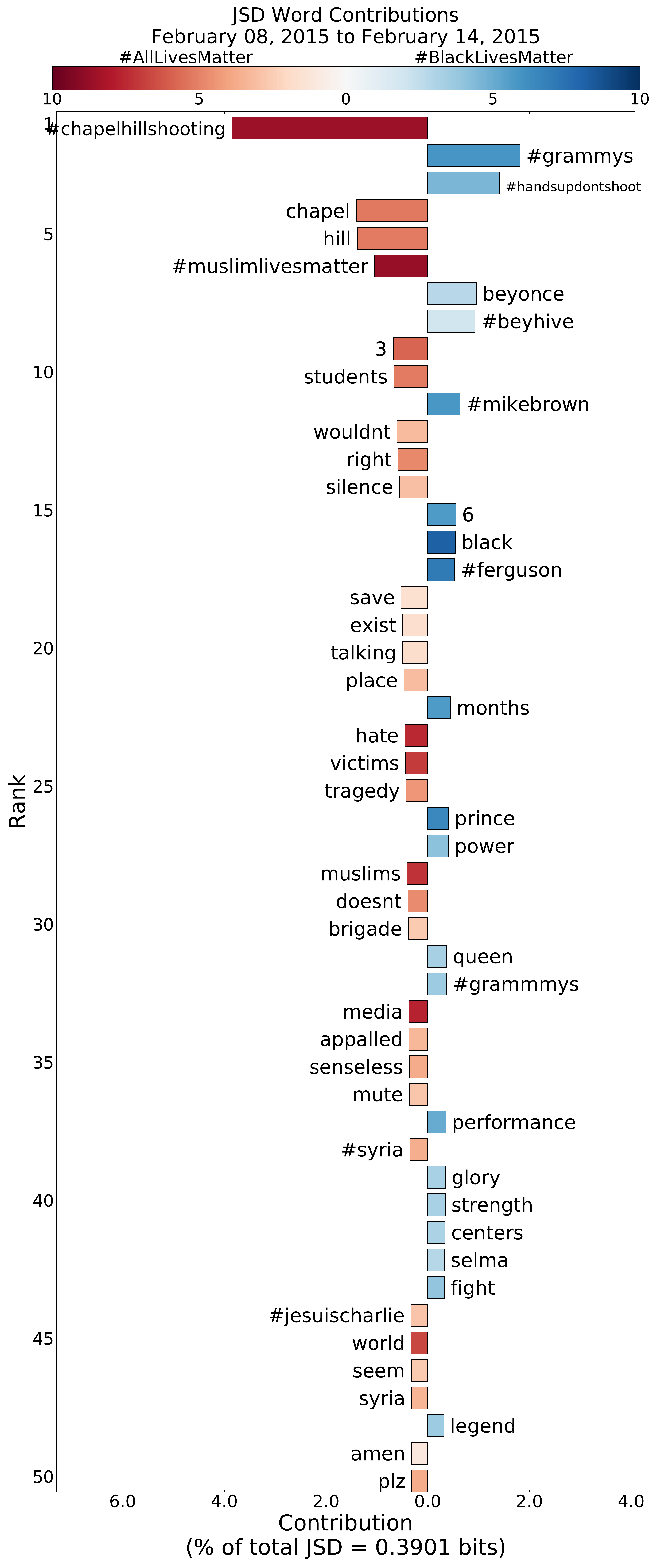}
\caption{Jensen-Shannon divergence word shift graph for the week following the 2015 Grammy Awards and the Chapel Hill shooting.  See main text and the caption of  Fig.~\ref{2014-11-24-wordshift} for an explanation of the word shift graphs. This period reflects a time in which usage of both \#BlackLivesMatter and \#AllLivesMatter diverged due to focus on different events. Discussion within \#AllLivesMatter reflects the Chapel Hill shooting, while discussion within \#BlackLivesMatter reflects the 2015 Grammy Awards and the performances of Beyonce and John Legend that alluded to Black Lives Matter.}
\label{2015-02-08-wordshift}
\end{minipage}
\quad
\end{figure*}

\begin{figure}
\includegraphics[scale = .215]{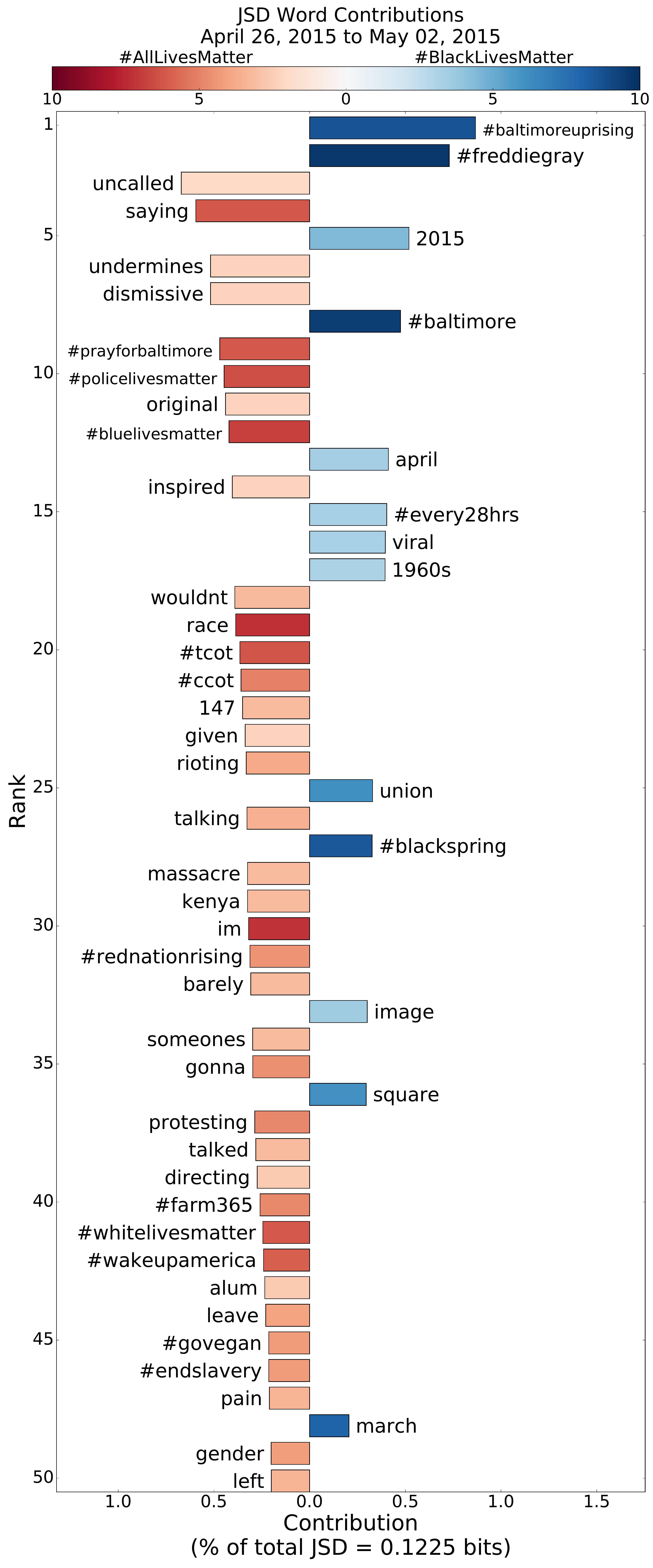}
\caption{Jensen-Shannon divergence word shift graph for the week encapsulating the peak of the Baltimore protests surrounding the death of Freddie Gray.  See main text and the caption of  Fig.~\ref{2014-11-24-wordshift} for an explanation of the word shift graphs. In this period, the conservative alignment of \#AllLivesMatter is particularly prevalent, as seen in the hashtags \#tcot (Top Conservatives on Twitter), \#ccot (Conservative Christians on Twitter), and \#rednationrising.}
\label{2015-04-26-wordshift}
\end{figure}


All contributions on the JSD word shift graphs are positive, where a bar to the left indicates the word was more common in \#AllLivesMatter and a bar to the right indicates the word was more common in \#BlackLivesMatter. The bars of the JSD word shift graph are also shaded according to the diversity of language surrounding each word. For each word $w$, we consider all tweets containing $w$ in the given hashtag. From these tweets, we calculate the Shannon index of the underlying word distribution with the word $w$ and hashtag removed. A high Shannon index indicates a high diversity of words which, in the context of Twitter, implies that the word $w$ was used in a variety of different tweets. On the other hand, a low Shannon index indicates that the word $w$ originates from a few popular retweets. We emphasize that here we are using the Shannon index not to compare diversities between words, but to simply determine if a word was used diversely or not. By using Figure~\ref{2015-04-04-wordshift}
as a baseline (a period where \#AllLivesMatter was dominated by one retweet), we determine a rule of thumb that a word is not used diversely if its Shannon index is less than approximately 3 bits.

By inspection of Figure~\ref{2014-11-24-wordshift}, we find that \#ferguson and \#fergusondecision, both hashtags relevant to the non-indictment of Darren Wilson for the death of Michael Brown, contribute to the divergence of \#BlackLivesMatter from \#AllLivesMatter. Similarly, in Figure~\ref{2015-04-26-wordshift} \#freddiegray emerges as a divergent hashtag during the Baltimore protests due to \#BlackLivesMatter. In each of these periods, \#BlackLivesMatter diverges from \#AllLivesMatter by talking proportionally more about the relevant deaths of Black Americans. Similar divergences appear in the other periods as well, as evidenced by the appearance of \#ericgarner, \#walterscott, and \#sandrabland in the respective Appendix word shift graphs.

During important protest periods, the conversation within \#AllLivesMatter diversifies itself around the lives of law enforcement officers. As shown in Figure~\ref{2015-04-26-wordshift}, during the Baltimore protests in which \#baltimoreuprising and \#baltimore were used significantly in \#BlackLivesMatter, users of \#AllLivesMatter responded with diverse usage of \#policelivesmatter and \#bluelivesmatter. Similarly, Figure~\ref{2014-12-20-wordshift}
shows that upon the death of the two NYPD officers, words such as ``officers,'' ``ramos,'' ``liu,'' and ``prayers'' appeared in a variety of \#AllLivesMatter tweets. In addition, pro-law enforcement hashtags such as \#policelivesmatter, \#nypd, \#nypdlivesmatter, and \#bluelivesmatter all contribute to the divergence of \#AllLivesMatter from \#BlackLivesMatter. Such divergence comes at the same time that the hashtags \#moa and \#blackxmas and words ``protest,'' ``mall,'' and ``america'' were prevalent in \#BlackLivesMatter due to Christmas protests, specifically at the Mall of America in Bloomington, Minnesota. So, in the midst of political protests by Black Lives Matter advocates, we see a law-enforcement-aligned response from \#AllLivesMatter.

During the period following the non-indictment of Darren Wilson, there are some words, such as ``oppression,'' ``structural,'' and ``brutality,'' that seem to suggest engagement from \#AllLivesMatter with the issues being discussed within \#BlackLivesMatter,  such as structural racism and police brutality. Since the diversities of these words are low, we can inspect popular retweets containing these words to understand how they were used. Doing so, we find that the words actually emerge in \#AllLivesMatter due to hijacking \cite{jackson2015hijacking}, the adoption of a hashtag to mock or criticize it. That is, these words appear not because of discussion of structural oppression and police brutality by \#AllLivesMatter advocates, but because \#BlackLivesMatter supporters are specifically critiquing the fact such discussions are not occurring within \#AllLivesMatter. (We have chosen to not provide direct references to these tweets so as to protect the identity of the original tweeter.) Similarly, a ``3-panel comic'' strip criticizing the notion of ``All Lives Matter'' circulated through \#AllLivesMatter following the death of Eric Garner (Figure~\ref{2014-12-03-wordshift}),
and after the Chapel Hill and Charleston Church shootings, \#BlackLivesMatter proponents leveraged \#AllLivesMatter to question why believers of the phrase were not more vocal (Figures~\ref{2015-02-08-wordshift} and \ref{2015-06-17-wordshift}).
We note that we are able to pick up on these instances of hijacking by inspecting words with high divergence, but low diversity (meaning the divergence comes almost entirely from the few retweets containing the word). This hijacking drives the divergence of \#AllLivesMatter from \#BlackLivesMatter in many of these periods.


\subsection{Topic Networks}
\label{topics}

Having analyzed the micro-level dynamics of word usage within \#BlackLivesMatter and \#AllLivesMatter, we turn to focus on the broader topics of these movements and how these topics coincide with the word-level analysis. Previous work on political polarization has used hashtags as a proxy for topics \cite{conover2011political, bode2015candidate, borge2015content, smith2014mapping, romero2011differences} and here we use the same interpretation. However, not all hashtags assist in understanding the broad topics. For example, \#retweet and \#lol are two such hashtags that frequently appear in tweets, but they provide no evidently relevant information about the events that are being discussed. Thus, we require a way of uncovering the most important topics and how they connect to one another.

To find these topics, we first construct hashtag networks for each of \#BlackLivesMatter and \#AllLivesMatter, where nodes are hashtags and weighted edges denote co-occurrence of these hashtags. To extract the core of this hashtag network, we apply the disparity filter, a method introduced by Serrano et al. that yields the ``multiscale backbone'' of a weighted network \cite{serrano2009extracting}. This backbone extracts statistically significant edges compared to a null model where all weights are uniformly distributed.
We take the topic network to be the largest connected component of the backbone. For significance level $\alpha < 0.03$, the disparity filter begins to force drastic drops in the number of nodes removed from the original hashtag network, as shown in Figure~\ref{significance-level}.
For this reason, we analyze the backbones only for $\alpha \geq 0.03$.

\begin{table}[t!]
\scriptsize
\centering
\begin{tabular}{l|c|c|c|c}
\textbf{\#BlackLivesMatter}     & Nodes                 & \% Original Nodes     & Edges                 & Clustering           \\ \hline \hline
Nov. 24--Nov. 30, 2014 & 243                   & 7.39\%                & 467                   & 0.0605               \\ \hline
Dec. 3--Dec. 9, 2014   & 339                   & 5.98\%                & 794                   & 0.0691               \\ \hline
Dec. 20--Dec. 26, 2014 & 187                   & 5.96\%                & 391                   & 0.1635               \\ \hline
Feb. 8--Feb. 14, 2015  & 70                    & 4.75\%                & 94                    & 0.1740               \\ \hline
Apr. 4--Apr. 10, 2015  & 80                    & 4.12\%                & 114                   & 0.1019               \\ \hline
Apr. 26--May 2, 2015   & 234                   & 5.54\%                & 471                   & 0.1068               \\ \hline
Jun. 17--Jun. 23, 2015 & 167                   & 6.35\%                & 246                   & 0.0746               \\ \hline
Jul. 21--Jul. 27, 2015 & 216                   & 6.18\%                & 393                   & 0.0914               \\ \hline
\textbf{\#AllLivesMatter }      & \multicolumn{1}{l|}{} & \multicolumn{1}{l|}{} & \multicolumn{1}{l|}{} & \multicolumn{1}{l}{} \\ \hline \hline
Nov. 24--Nov. 30, 2014 & 26                    & 5.76\%                & 35                    & 0.1209               \\ \hline
Dec. 3--Dec. 9, 2014   & 31                    & 3.92\%                & 49                    & 0.2667               \\ \hline
Dec. 20--Dec. 26, 2014 & 41                    & 3.95\%                & 70                    & 0.2910               \\ \hline
Feb. 8--Feb. 14, 2015  & 18                    & 3.50\%                & 23                    & 0.1894               \\ \hline
Apr. 4--Apr. 10, 2015  & 7                     & 1.88\%                & 6                     & 0.0000               \\ \hline
Apr. 26--May 2, 2015   & 38                    & 4.12\%                & 62                    & 0.1868               \\ \hline
Jun. 17--Jun. 23, 2015 & 22                    & 4.56\%                & 28                    & 0.3571               \\ \hline
Jul. 21--Jul. 27, 2015 & 33                    & 4.26\%                & 44                    & 0.1209              
\end{tabular}
\caption{Summary statistics for topic networks created from the full hashtag networks using the disparity filter at the significance level $\protect\alpha$ = 0.03 \cite{serrano2009extracting}.}
\label{backbone-stats-alpha3}
\end{table}

\begin{figure*}
\centering
\includegraphics[scale=.7, trim = 50 0 0 0]{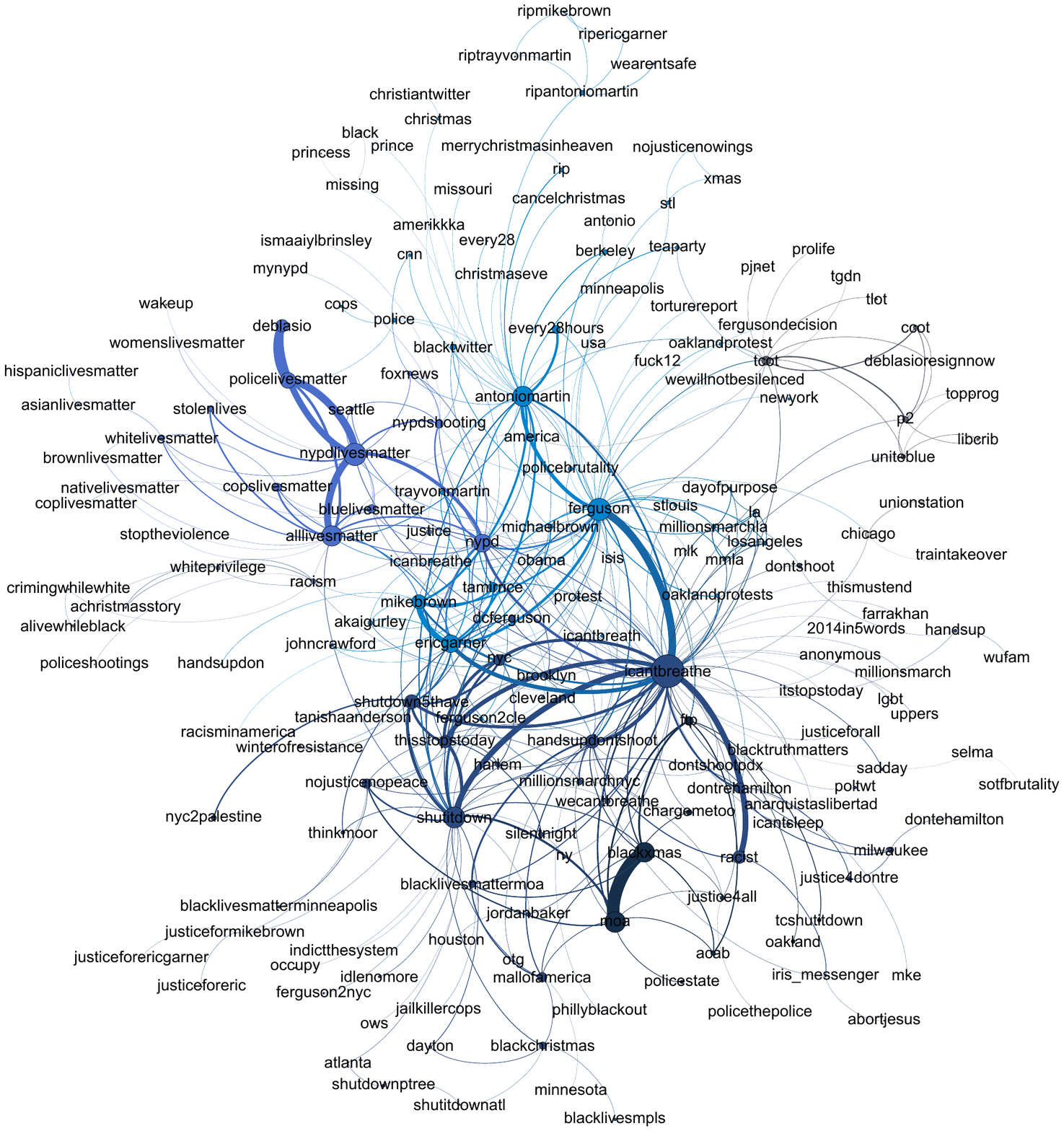}
\caption{\#BlackLivesMatter topic network for the week following the death of two NYPD officers.}
\label{2014-12-20-backbone-blacklives}
\end{figure*}

Visualizations of the topic networks following the week of the death of the two NYPD officers are presented in Figures~\ref{2014-12-20-backbone-blacklives} and \ref{2014-12-20-backbone-alllives}, and networks of the remaining periods are shown in the Appendix.
Node sizes are proportional to the square root of the number of times each hashtag was used, and node colors are determined by the Louvain structure detection method \cite{blondel2008fast}. The exact assignments of topics to communities is not critical, but rather they provide a visual guide through the networks.

\begin{figure*}
\includegraphics[scale=.3]{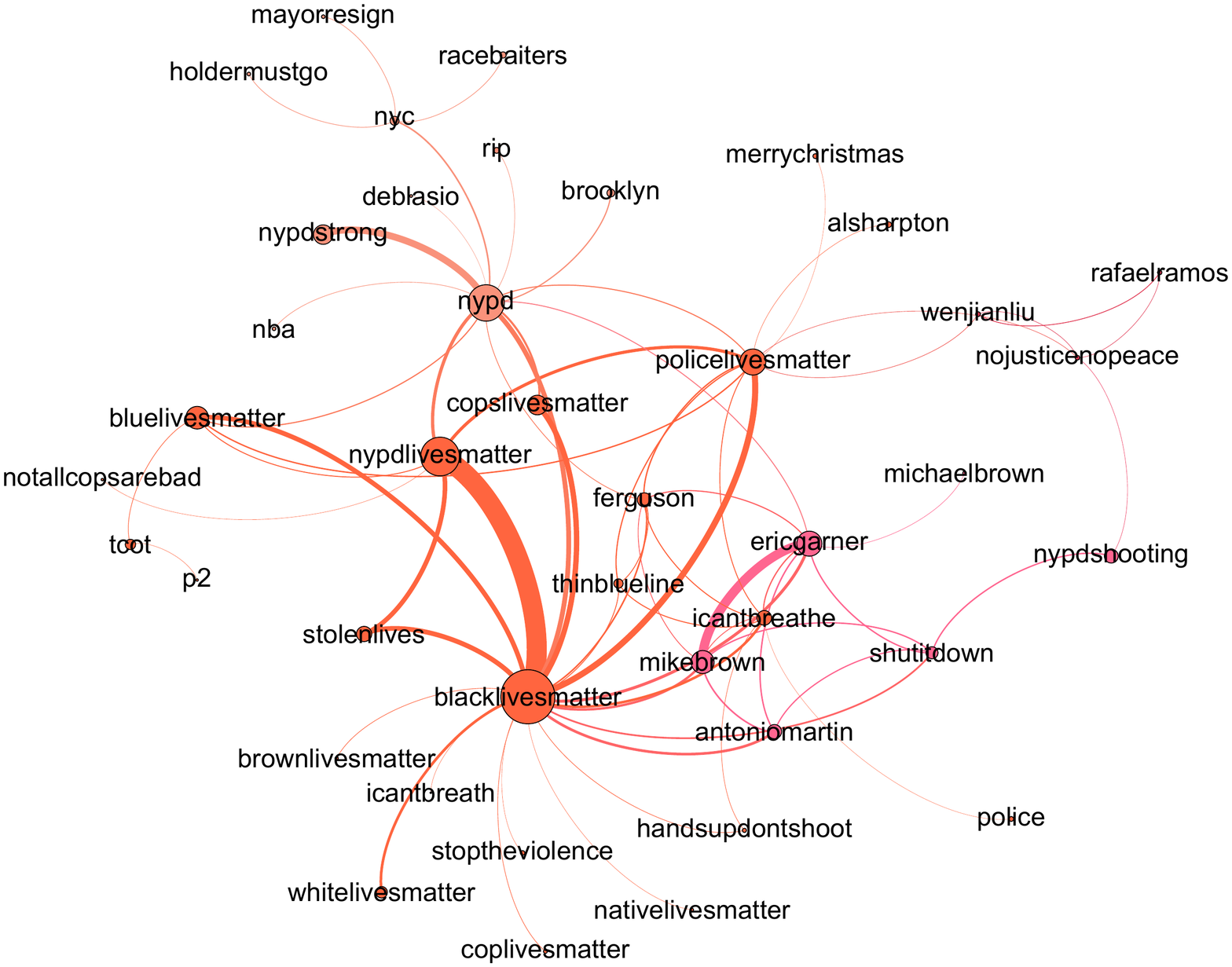}
\caption{\#AllLivesMatter topic network for the week following the death of two NYPD officers.}
\label{2014-12-20-backbone-alllives}
\end{figure*}

\begin{table*}[]
\scriptsize
\centering
\begin{tabular}{ll|c|l|c}
                                            & \multicolumn{2}{c|}{\textbf{\#BlackLivesMatter}}                                                                                                                                                                                                                                                                                                                          & \multicolumn{2}{c}{\textbf{\#AllLivesMatter}}                                                                                                                                                                                                                                                                                                                                      \\
\multicolumn{1}{l|}{}                       & \multicolumn{1}{c|}{Top Hashtags}                                                                                                                                                                                                       & Betweenness                                                                                                                     & \multicolumn{1}{c|}{Top Hashtags}                                                                                                                                                                                                                & \multicolumn{1}{c}{Betweenness}                                                                                                \\ \hline \hline
\multicolumn{1}{l|}{Nov. 24--Nov. 30, 2014} & \begin{tabular}[c]{@{}l@{}}1. ferguson\\ 2. mikebrown\\ 3. fergusondecision\\ 4. shutitdown\\ 5. justiceformikebrown\\ 6. tamirrice\\ 7. blackoutblackfriday\\ 8. alllivesmatter\\ 9. boycottblackfriday\\ 10. blackfriday\end{tabular} & \begin{tabular}[c]{@{}c@{}}0.8969\\ 0.2303\\ 0.1159\\ 0.0854\\ 0.0652\\ 0.0583\\ 0.0577\\ 0.0512\\ 0.0442\\ 0.0439\end{tabular} & \begin{tabular}[c]{@{}l@{}}1. blacklivesmatter\\ 2. ferguson\\ 3. fergusondecision\\ 4. mikebrown\\ 5. nycprotest\\ 6. williamsburg\\ 7. brownlivesmatter\\ 8. sf\\ 9. blackfridayblackout\\ 10. nyc\end{tabular}                                & \begin{tabular}[c]{@{}c@{}}0.7229\\ 0.6079\\ 0.1791\\ 0.1730\\ 0.0303\\ 0.0192\\ 0.0158\\ 0.0150\\ 0.0137\\ 0.0123\end{tabular} \\ \hline
\multicolumn{1}{l|}{Dec. 3--Dec. 9, 2014}   & \begin{tabular}[c]{@{}l@{}}1. ericgarner\\ 2. icantbreathe\\ 3. ferguson\\ 4. shutitdown\\ 5. mikebrown\\ 6. thisstopstoday\\ 7. handsupdontshoot\\ 8. nojusticenopeace\\ 9. nypd\\ 10. berkeley\end{tabular}                           & \begin{tabular}[c]{@{}c@{}}0.6221\\ 0.5035\\ 0.2823\\ 0.1117\\ 0.0745\\ 0.0505\\ 0.0469\\ 0.0449\\ 0.0399\\ 0.0352\end{tabular} & \begin{tabular}[c]{@{}l@{}}1. blacklivesmatter\\ 2. ericgarner\\ 3. ferguson\\ 4. icantbreathe\\ 5. tcot\\ 6. shutitdown\\ 7. handsupdontshoot\\ 8. crimingwhilewhite\\ 9. miami\\ 10. mikebrown\end{tabular}                                    & \begin{tabular}[c]{@{}c@{}}0.6589\\ 0.4396\\ 0.3684\\ 0.2743\\ 0.1916\\ 0.1529\\ 0.0797\\ 0.0666\\ 0.0666\\ 0.0645\end{tabular} \\ \hline
\multicolumn{1}{l|}{Dec. 20--Dec. 26, 2014} & \begin{tabular}[c]{@{}l@{}}1. icantbreathe\\ 2. ferguson\\ 3. antoniomartin\\ 4. shutitdown\\ 5. nypd\\ 6. alllivesmatter\\ 7. ericgarner\\ 8. nypdlivesmatter\\ 9. tcot\\ 10. handsupdontshoot\end{tabular}                            & \begin{tabular}[c]{@{}c@{}}0.5742\\ 0.3234\\ 0.2826\\ 0.2473\\ 0.1496\\ 0.1324\\ 0.1265\\ 0.1147\\ 0.1082\\ 0.0986\end{tabular} & \begin{tabular}[c]{@{}l@{}}1. blacklivesmatter\\ 2. nypd\\ 3. policelivesmatter\\ 4. nypdlivesmatter\\ 5. ericgarner\\ 6. nyc\\ 7. bluelivesmatter\\ 8. icantbreathe\\ 9. shutitdown\\ 10. mikebrown\end{tabular}                                & \begin{tabular}[c]{@{}c@{}}0.6791\\ 0.4486\\ 0.2926\\ 0.1990\\ 0.1565\\ 0.1461\\ 0.1409\\ 0.1254\\ 0.0992\\ 0.0860\end{tabular} \\ \hline
\multicolumn{1}{l|}{Feb. 8--Feb. 14, 2015}  & \begin{tabular}[c]{@{}l@{}}1. blackhistorymonth\\ 2. grammys\\ 3. muslimlivesmatter\\ 4. bhm\\ 5. alllivesmatter\\ 6. handsupdontshoot\\ 7. mikebrown\\ 8. ferguson\\ 9. icantbreathe\\ 10. beyhive\end{tabular}                        & \begin{tabular}[c]{@{}c@{}}0.6506\\ 0.5614\\ 0.5515\\ 0.4987\\ 0.4932\\ 0.4150\\ 0.2588\\ 0.1754\\ 0.1614\\ 0.1325\end{tabular} & \begin{tabular}[c]{@{}l@{}}1. muslimlivesmatter\\ 2. blacklivesmatter\\ 3. chapelhillshooting\\ 4. jewishlivesmatter\\ 5. butinacosmicsensenothingreallymatters\\ 6. whitelivesmatter\\ 7. rip\\ 8. hatecrime\\ 9. ourthreewinners\end{tabular} & \begin{tabular}[c]{@{}c@{}}0.8012\\ 0.4296\\ 0.4192\\ 0.2279\\ 0.0431\\ 0.0112\\ 0.0073\\ 0.0071\\ 0.0058\end{tabular}          \\ \hline
\multicolumn{1}{l|}{Apr. 4--Apr. 10, 2015}  & \begin{tabular}[c]{@{}l@{}}1. walterscott\\ 2. blacktwitter\\ 3. icantbreathe\\ 4. ferguson\\ 5. p2\\ 6. ericgarner\\ 7. alllivesmatter\\ 8. mlk\\ 9. kendrickjohnson\\ 10. tcot\end{tabular}                                           & \begin{tabular}[c]{@{}c@{}}0.9118\\ 0.2283\\ 0.1779\\ 0.1555\\ 0.1022\\ 0.1003\\ 0.0906\\ 0.0717\\ 0.0685\\ 0.0617\end{tabular} & 1. blacklivesmatter                                                                                                                                                                                                                              & 1.0000                                                                                                                          \\ \hline
\multicolumn{1}{l|}{Apr. 26--May 2, 2015}   & \begin{tabular}[c]{@{}l@{}}1. freddiegray\\ 2. baltimore\\ 3. baltimoreuprising\\ 4. baltimoreriots\\ 5. alllivesmatter\\ 6. mayday\\ 7. blackspring\\ 8. tcot\\ 9. baltimoreuprising\\ 10. handsupdontshoot\end{tabular}               & \begin{tabular}[c]{@{}c@{}}0.5806\\ 0.4732\\ 0.2625\\ 0.2415\\ 0.1053\\ 0.0970\\ 0.0737\\ 0.0581\\ 0.0500\\ 0.0458\end{tabular} & \begin{tabular}[c]{@{}l@{}}1. blacklivesmatter\\ 2. baltimoreriots\\ 3. baltimore\\ 4. freddiegray\\ 5. policelivesmatter\\ 6. baltimoreuprising\\ 7. tcot\\ 8. peace\\ 9. whitelivesmatter\\ 10. wakeupamerica\end{tabular}                     & \begin{tabular}[c]{@{}c@{}}0.7227\\ 0.4339\\ 0.3463\\ 0.1869\\ 0.1106\\ 0.1014\\ 0.0663\\ 0.0451\\ 0.0444\\ 0.0281\end{tabular} \\ \hline
\multicolumn{1}{l|}{Jun. 17--Jun. 23, 2015} & \begin{tabular}[c]{@{}l@{}}1. charlestonshooting\\ 2. charleston\\ 3. blacktwitter\\ 4. tcot\\ 5. unitedblue\\ 6. racism\\ 7. ferguson\\ 8. usa\\ 9. takedowntheflag\\ 10. baltimore\end{tabular}                                       & \begin{tabular}[c]{@{}c@{}}0.8849\\ 0.2551\\ 0.1489\\ 0.1379\\ 0.1340\\ 0.1323\\ 0.1085\\ 0.1011\\ 0.0790\\ 0.0788\end{tabular} & \begin{tabular}[c]{@{}l@{}}1. charlestonshooting\\ 2. blacklivesmatter\\ 3. bluelivesmatter\\ 4. gunsense\\ 5. pjnet\\ 6. 2a\\ 7. wakeupamerica\\ 8. tcot\\ 9. gohomederay\\ 10. ferguson\end{tabular}                                           & \begin{tabular}[c]{@{}c@{}}0.6666\\ 0.6238\\ 0.4900\\ 0.2571\\ 0.2292\\ 0.1857\\ 0.1494\\ 0.1289\\ 0.0952\\ 0.0952\end{tabular} \\ \hline
\multicolumn{1}{l|}{Jul. 21--Jul. 27, 2015} & \begin{tabular}[c]{@{}l@{}}1. sandrabland\\ 2. sayhername\\ 3. justiceforsandrabland\\ 4. unitedblue\\ 5. blacktwitter\\ 6. alllivesmatter\\ 7. tcot\\ 8. defundpp\\ 9. p2\\ 10. m4bl\end{tabular}                                      & \begin{tabular}[c]{@{}c@{}}0.7802\\ 0.3175\\ 0.1994\\ 0.1870\\ 0.1788\\ 0.1648\\ 0.1081\\ 0.0827\\ 0.0756\\ 0.0734\end{tabular} & \begin{tabular}[c]{@{}l@{}}1. blacklivesmatter\\ 2. pjnet\\ 3. tcot\\ 4. uniteblue\\ 5. defundplannedparenthood\\ 6. defundpp\\ 7. sandrabland\\ 8. justiceforsandrabland\\ 9. prolife\\ 10. nn15\end{tabular}                                   & \begin{tabular}[c]{@{}c@{}}0.8404\\ 0.2689\\ 0.2683\\ 0.2440\\ 0.2437\\ 0.1692\\ 0.1386\\ 0.1386\\ 0.0881\\ 0.0625\end{tabular}
\end{tabular}
\caption{The top 10 hashtags in the topic networks as determined by random walk betweeness centrality for each time period. Some \#AllLivesMatter topic networks have less than 10 top nodes due to the relatively small size of the networks.}
\label{random-walk-betweenness}
\end{table*}

Table \ref{backbone-stats-alpha3} and Appendix Tables~\ref{backbone-stats-alpha4} and \ref{backbone-stats-alpha5}
report the number of nodes, edges, clustering coefficients, and percentages of nodes maintained from the full hashtag networks per each significance level $\alpha = 0.03, 0.04$, and $0.05$ respectively. We see that across all time periods of interest and significance levels, the number of topics in \#BlackLivesMatter is higher than that of \#AllLivesMatter. We also note that the clustering of the \#BlackLivesMatter topics is less that of \#AllLivesMatter almost always. Thus, not only are there more topics presented in \#BlackLivesMatter, but they are more diverse in their connections. In contrast, the stronger \#AllLivesMatter ties, as measured by their clustering, suggest that the \#AllLivesMatter topics are more tightly connected and revolve around similar themes. We see the clustering is less within \#AllLivesMatter during the week of Walter Scott's death, where the topic network has a star-like shape with no triadic closure across all significance levels. This low clustering is not indicative of diverse conversation, as the central node \#BlackLivesMatter connects several disparate topics. This is in line with our word-level analysis, where we concluded from Figure~\ref{2015-04-04-wordshift}
that the discussion within \#AllLivesMatter was dominated by a retweet not pertaining to the death of Walter Scott, the event of that time period.

In order to extract the most central topics of \#BlackLivesMatter and \#AllLivesMatter during each time period, we compare the results of three centrality measures, betweenness centrality \cite{newman2004finding}, random walk betweeness centrality \cite{girvan2002community}, and PageRank \cite{page1999pagerank} on the topic networks at significance level $\alpha = 0.03$. Through inspection of the rankings of each list, we find the relative rankings of the most central topics in \#BlackLivesMatter and \#AllLivesMatter are robust to the centrality measure used. Table~\ref{random-walk-betweenness} shows the rankings according to random walk centrality. Rankings from betweenness centrality and PageRank are reported in Appendix Tables~\ref{backbone-betweenness} and \ref{pagerank}.

\begin{figure*}[!htb]
\includegraphics[scale=.4]{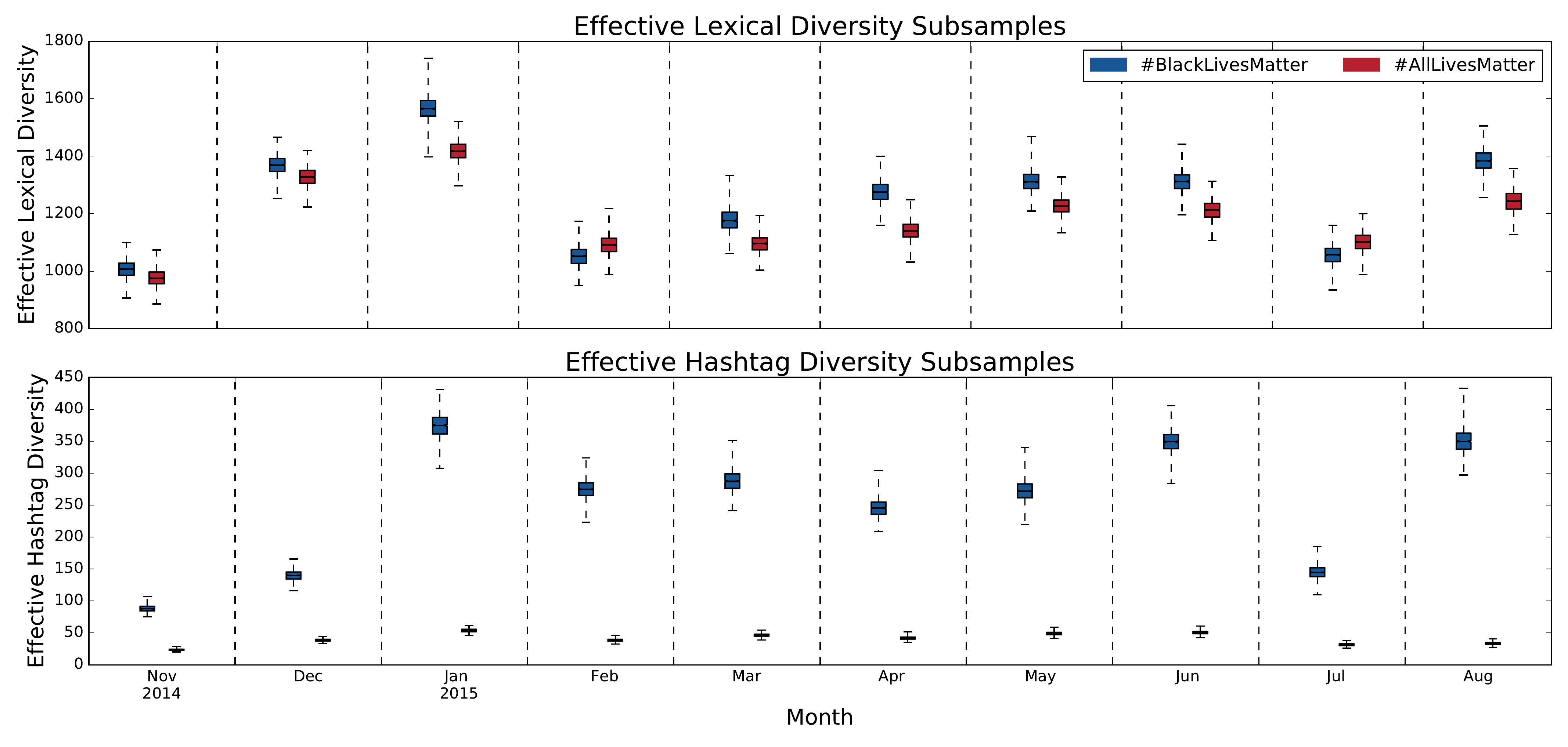}
\caption{To control for how volume affects the effective diversity of \#BlackLivesMatter and \#AllLivesMatter, we break the time scale down into months and subsample 2,000 tweets from each hashtag 1,000 times. The plot shows notched box plots depicting the distributions of these subsamples for effective lexical and hashtag diversity. The notches are small on all the boxes, indicating that the mean diversities are significantly different at the 95\% confidence level across all time periods.}
\label{conversation-diversity-subsample}
\end{figure*}

We see that for both \#BlackLivesMatter and \#AllLivesMatter, the top identified topics are indicative of the relative events occurring in each time period. In \#BlackLivesMatter for instance, \#ferguson and \#mikebrown are top topics after the non-indictment of Darren Wilson, \#walterscott is a top topic after the death of Walter Scott, and \#sandrabland and \#sayhername are a top topics during the time period following the death of Sandra Bland. However, these major topics rank differently in both \#BlackLivesMatter and \#AllLivesMatter. For instance, while \#mikebrown, \#ericgarner, \#icantbreathe, \#freddiegray, \#baltimore, and \#sandrabland all consistently rank higher in \#BlackLivesMatter than in \#AllLivesMatter. The significant presence of these hashtags within \#BlackLivesMatter is consist with our findings from the JSD word shift graphs.

The most prominent discussion of non-Black lives in the topic networks of \#AllLivesMatter is discussion of police lives.  We see that in \#AllLivesMatter, \#nypd, \#policelivesmatter, and \#bluelivesmatter are ranked higher as topics in \#AllLivesMatter than in \#BlackLivesMatter during December 20th and April 26th periods, similar to what we found in the JSD word shift graphs. On the other hand, hashtags depicting strong anti-police sentiment such as \#killercops, \#policestate, and \#fuckthepolice appear almost exclusively in \#BlackLivesMatter and are absent from \#AllLivesMatter. The alignment of \#AllLivesMatter with police lives coincides with a broader alignment with the conservative sphere of Twitter that is apparent through the topic networks. In several periods for \#AllLivesMatter, \#tcot is a central topic, as well as \#pjnet (Patriots Journal Network), \#wakeupamerica, and \#defundplannedparenthood. The hashtag \#tcot also appears in several of the \#BlackLivesMatter periods as well. This is to be expected, as Freelon et al. found that a portion of \#BlackLivesMatter tweets were hijacked by the conservative sphere of Twitter \cite{freelon2016beyond}.

However, the hijacking of \#BlackLivesMatter and content injection of conservative views is a much smaller component of the \#BlackLivesMatter topics as compared to the respective hijacking of the \#AllLivesMatter topics. As evidenced both by the network statistics and the network visualizations themselves, the \#BlackLivesMatter topic networks show that the conversations are diverse and multifaceted while the \#AllLivesMatter networks show conversations that are more limited in scope. Furthermore, \#BlackLivesMatter is consistently a more central topic within the \#AllLivesMatter networks than \#AllLivesMatter is within the \#BlackLivesMatter networks. Thus, hijacking is more prevalent within \#AllLivesMatter, while \#BlackLivesMatter users are able to maintain diverse conversations and delegate hijacking to only a portion of the discourse.


\subsection{Conversational Diversity}
\label{conversational diversity}

Having quantified both the word-level divergences and the large-scale topic networks, we now measure the informational diversity of \#BlackLivesMatter and \#AllLivesMatter more precisely. We do this through two approaches. First, we measure ``lexical diversity,'' the diversity of words other than hashtags. Second, we measure the hashtag diversity. We measure these diversities using the effective diversity described in Eqn.~\ref{effective-diversity} in Section~\ref{entropy-and-diversity}. Furthermore, to account for the different volume of \#BlackLivesMatter and \#AllLivesMatter tweets, we break the time scale down into months and subsample 2,000 tweets from each hashtag 1,000 times, calculating the effective diversities each time. The results are shown in Figure~\ref{conversation-diversity-subsample}.

The lexical diversity of \#BlackLivesMatter is larger than \#AllLivesMatter in eight of the ten months with an average lexical diversity that is 5\% more than that of \#AllLivesMatter. Interestingly, the two cases where \#AllLivesMatter has higher lexical diversity are in the two periods when there were large non-police-involved shootings of people of color and \#AllLivesMatter was used as a hashtag of solidarity.  However, the more striking differences are in terms of hashtag diversity. On average, the hashtag diversity of \#BlackLivesMatter is six times that of \#AllLivesMatter. This is in line with our network analysis where we found expansive \#BlackLivesMatter topic networks and tightly clustered, less diverse \#AllLivesMatter topic networks.

The low hashtag diversity of \#AllLivesMatter is relatively constant. One could imagine that the lack of diversity in the topics of \#AllLivesMatter is a result of a focused conversation that does not deviate from its main message.  However, as we have demonstrated through the JSD word shift graphs and topic networks, the conversation of \#AllLivesMatter does change and evolve with respect to the different time periods. We see mentions of the major deaths and events of these periods within \#AllLivesMatter, even if they do not rank as highly in terms of topic centrality. So, even though both protest hashtags have overlap on many of the major topics, the diversity of topics found within \#BlackLivesMatter far exceeds that of \#AllLivesMatter, even when accounting for volume.

\section{Discussion.}

Through a multi-level analysis of \#BlackLivesMatter and \#AllLivesMatter tweets, ranging from the word level to the topic level, we have demonstrated key differences between the two protest groups. Although one of these differences has been proportionally higher discussion of Black deaths in \#BlackLivesMatter, it is important to note that \#AllLivesMatter is not completely devoid of discussion about these deaths. For instance, \#ripericgarner is prominent within \#AllLivesMatter following the death of Eric Garner, \#iamame (``I am African Methodist Episcopal'') contributes more to \#AllLivesMatter following the Charleston Church shooting, and the names of several Black Americans appear in the \#AllLivesMatter topic networks. However, many of these signs of solidarity are associated with low diversity. In light of this, it is also important to note that there is a lack of discussion of other deaths within \#AllLivesMatter. That is, in examining several of the main periods where \#AllLivesMatter spikes, only the Chapel Hill shooting period (Figures~\ref{2015-02-08-wordshift} and \ref{2015-02-08-backbone-alllives})
shows discussion of non-Black deaths.

The word shift graphs and topic networks reveal that the only other lives that are significantly discussed within \#AllLivesMatter are the lives of law enforcement officers, particularly during times in which there is heavy protesting. Although the notions of ``Black Lives Matter'' and ``Police Lives Matter'' are not necessarily mutually exclusive \cite{speight2015black}, we see that the conversations within \#AllLivesMatter often frame Black protesters versus law enforcement with an``us versus them'' mentality. This framing echoes the ways in which media outlets have historically framed the tension between Black protesters and law enforcement \cite{van1989race,van2000new,lackey2005framing}, where police officers and protesters are seen as ``enemy combatants'' \cite{rickford2016black} and such movements appear to ``jeopardize law enforcement lives'' \cite{speight2015black}. So, by facilitating this opposition, \#AllLivesMatter becomes the center of upholding historically contentious views in the midst of what some consider the New Civil Rights Movement.

In line with Freelon et al.'s findings, we have uncovered hijacking of \#BlackLivesMatter  by \#AllLivesMatter advocates \cite{freelon2016beyond}. Such hijacking is similar to the content injection described by Conover et al. \cite{conover2011political}, where one group adopts the hashtag of politically opposed group in order to inject their ideological beliefs. Such content injection has also between found in the work of Egyptian political polarization \cite{borge2015content}. However, a significant portion \#AllLivesMatter hijacking by \#BlackLivesMatter supporters is not simple content injection. Rather, advocates of \#BlackLivesMatter often use \#AllLivesMatter to directly interrogate the stance of ``All Lives Matter'' and the worldview implied by that phrase. Furthermore, such discussions have largely been relegated to \#AllLivesMatter, allowing \#BlackLivesMatter to exhibit diverse conversations about a variety of topics. Although past research has expressed concern that \#AllLivesMatter would derail from the movement started by \#BlackLivesMatter \cite{orbe2015black,rickford2016black,speight2015black,carney2016all}, our data-driven approach has allowed us to uncover that \#BlackLivesMatter has countered \#AllLivesMatter content injection.

We were largely able to unpack these narratives by looking not just as hashtag trends and curated term lists, as in previous studies, but at the entirety of the text and how it is structured at both the word and topic levels. In this sense, our work falls closer to that of researchers who have constructed networks of these protest movements to inform largely qualitative research \cite{papacharissi2012affective, jackson2015ferguson, jackson2015hijacking}, rather than research that focuses solely on the topological properties of the networks. Our methods generalize to studying how the discourse of two or more protest movements diverge from one another and how these differences can be captured at the word and topic levels.  Furthermore, by integrating diversities within our divergence analysis, we have shown how we can measure word-level differences between protests and counter-protests in such a way that facilitates follow-up qualitative investigation of how exactly those words contribute to the divergence. Overall, these tools provide an avenue for exploring more of the data narratives between \#BlackLivesMatter and \#AllLivesMatter, and other instances of political polarization.


\section*{Acknowledgements}

All authors would like to thank the members of the Computational Story Lab for their feedback and support. They would also like to thank James Bagrow for his helpful suggestions. PSD and CMD acknowledge support from NSF Big Data Grant \#1447634.


\bibliography{bibl}
\bibliographystyle{unsrtabbrv}


\appendix

\renewcommand\thefigure{\thesection\arabic{figure}}  
\renewcommand\thetable{\thesection\arabic{table}}

\setcounter{figure}{0} 
\setcounter{table}{0}

\begin{figure*}
\section{JSD Word Shift Graphs}
\begin{minipage}[t]{0.45\linewidth}
\includegraphics[scale = .215]{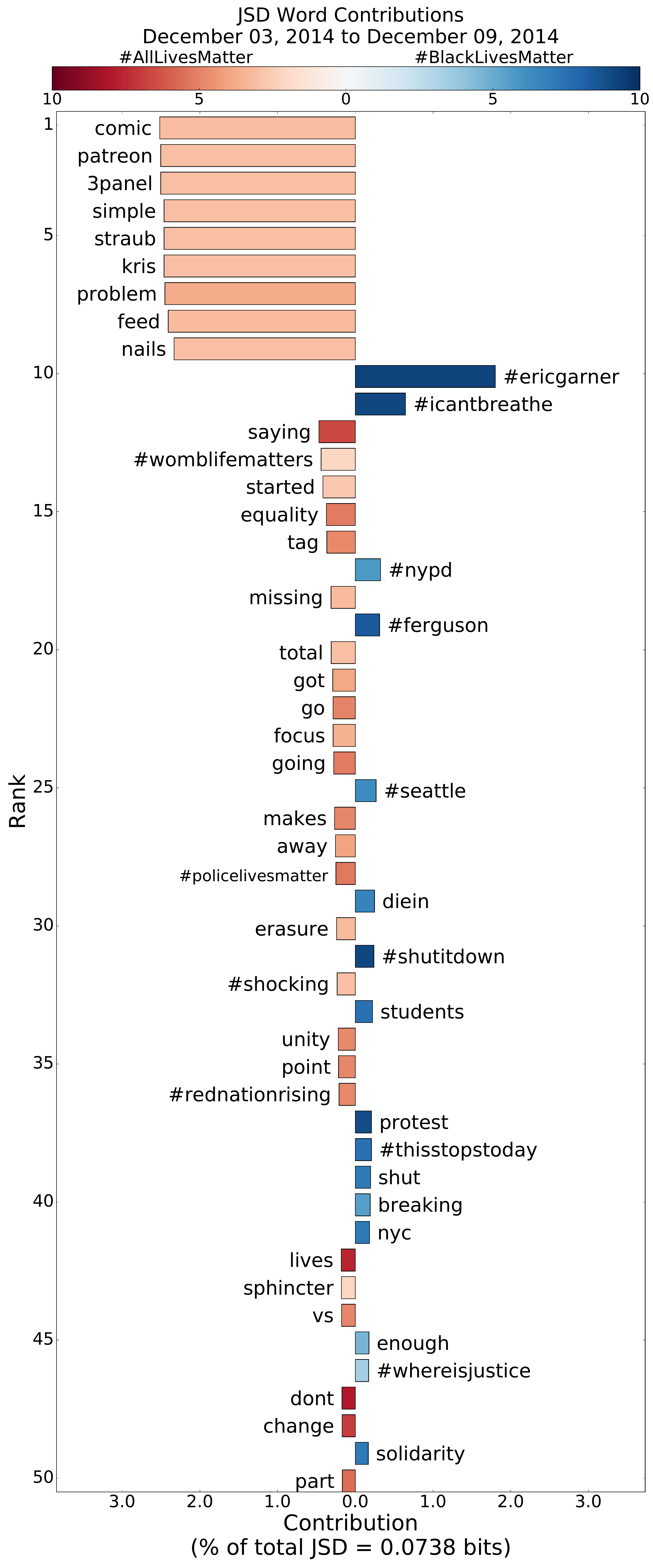}
\caption{Jensen-Shannon divergence word shift graph for the week following the non-indictment of Daniel Pantaleo. See main text and the caption of  Fig.~\ref{2014-11-24-wordshift} for an explanation of the word shift graphs. The leading contributors to the divergence on the side of \#AllLivesMatter are due to a popular retweet discussing a ``3-panel'' sketch ``comic'' that describes what the ``problem'' with \#AllLivesMatter is.}
\label{2014-12-03-wordshift}
\end{minipage}
\quad
\begin{minipage}[t]{0.45\linewidth}
\includegraphics[scale = .215]{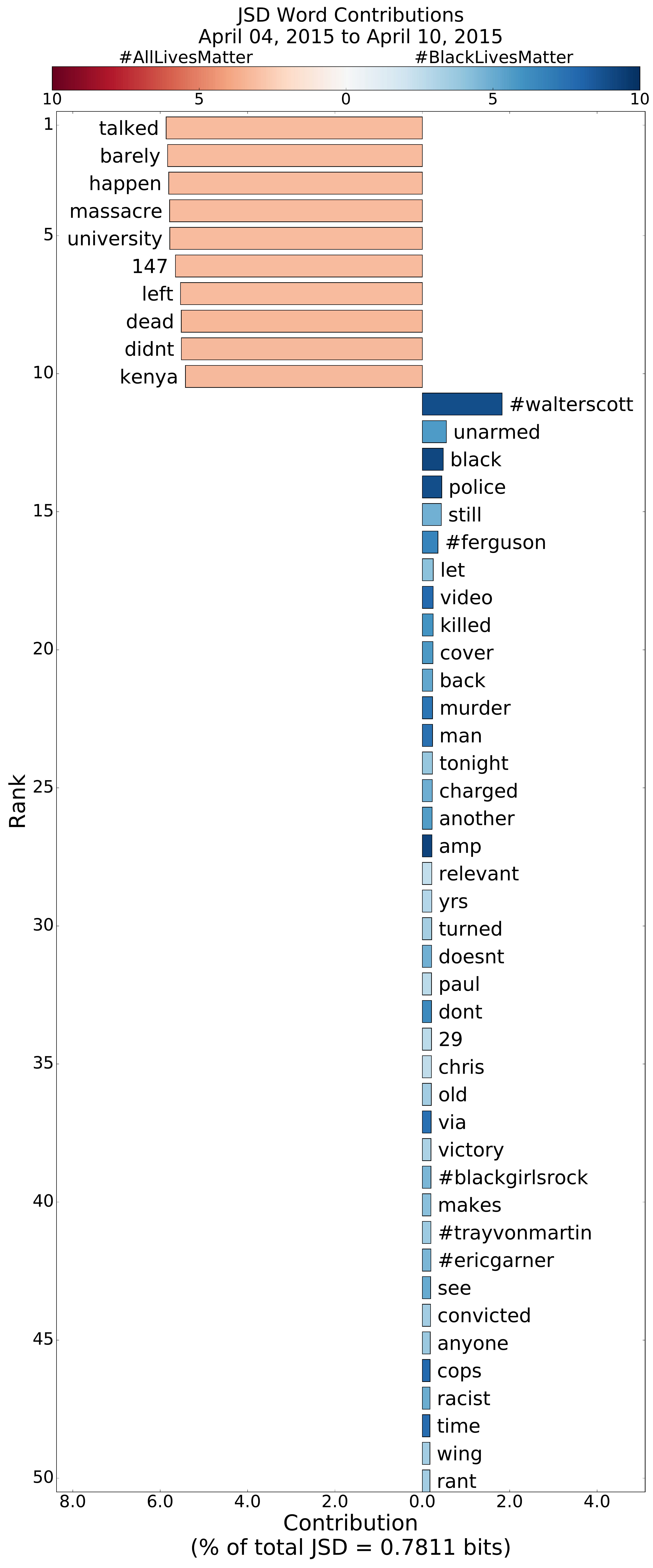}
\caption{Jensen-Shannon divergence word shift graph for the week following the death of Walter Scott. See main text and the caption of  Fig.~\ref{2014-11-24-wordshift} for an explanation of the word shift graphs. The leading contributors to the divergence on the side of \#AllLivesMatter are due to a popular retweet that asks why people are not paying attention to a school massacre that occurred in Kenya. This retweet is persistent, which is why the words ``kenya'' and ``massacre''  appear in later dates, such as in Fig. \ref{2015-04-26-wordshift}}.
\label{2015-04-04-wordshift}
\end{minipage}
\end{figure*}

\begin{figure*}
\begin{minipage}[t]{0.45\linewidth}
\includegraphics[scale = .215]{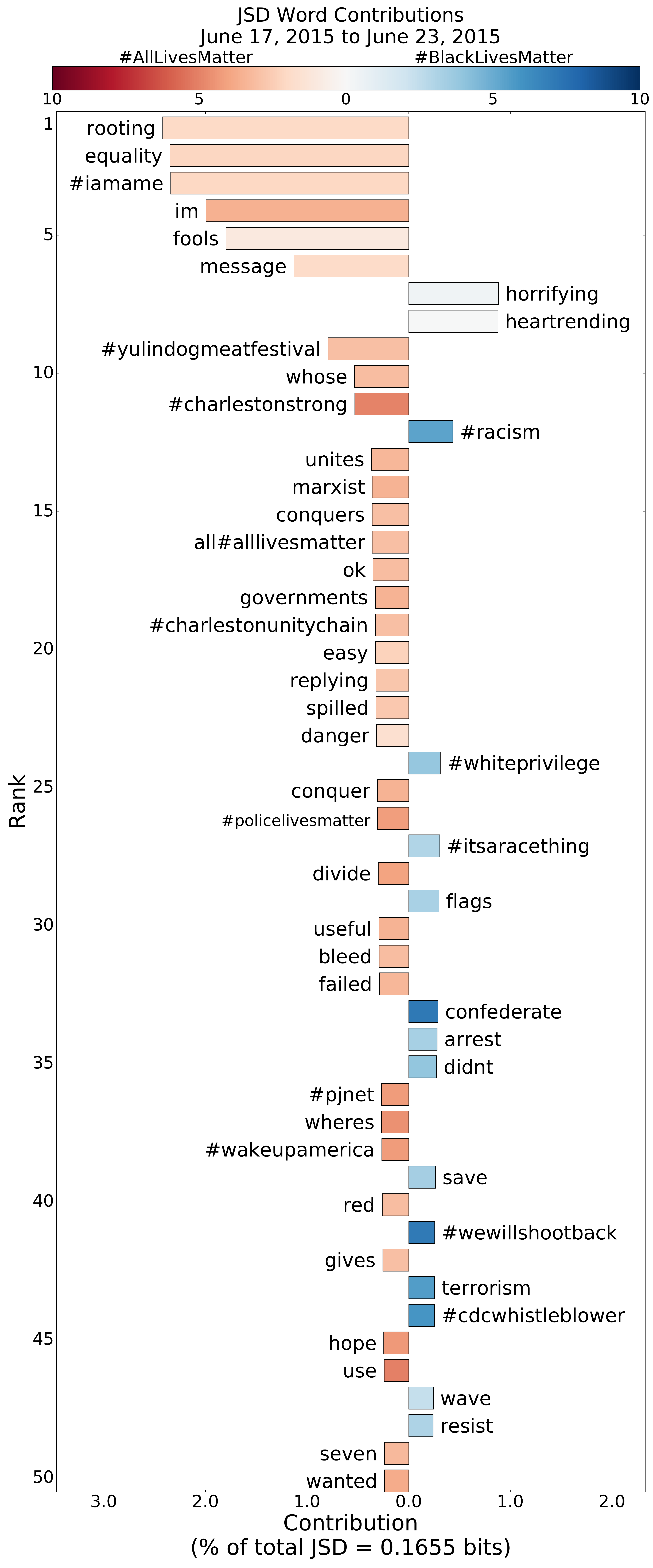}
\caption{Jensen-Shannon divergence word shift for the week following the Charleston church shooting. See main text and the caption of  Fig.~\ref{2014-11-24-wordshift} for an explanation of the word shift graphs.}
\label{2015-06-17-wordshift}
\end{minipage}
\quad
\begin{minipage}[t]{0.45\linewidth}
\includegraphics[scale = .215]{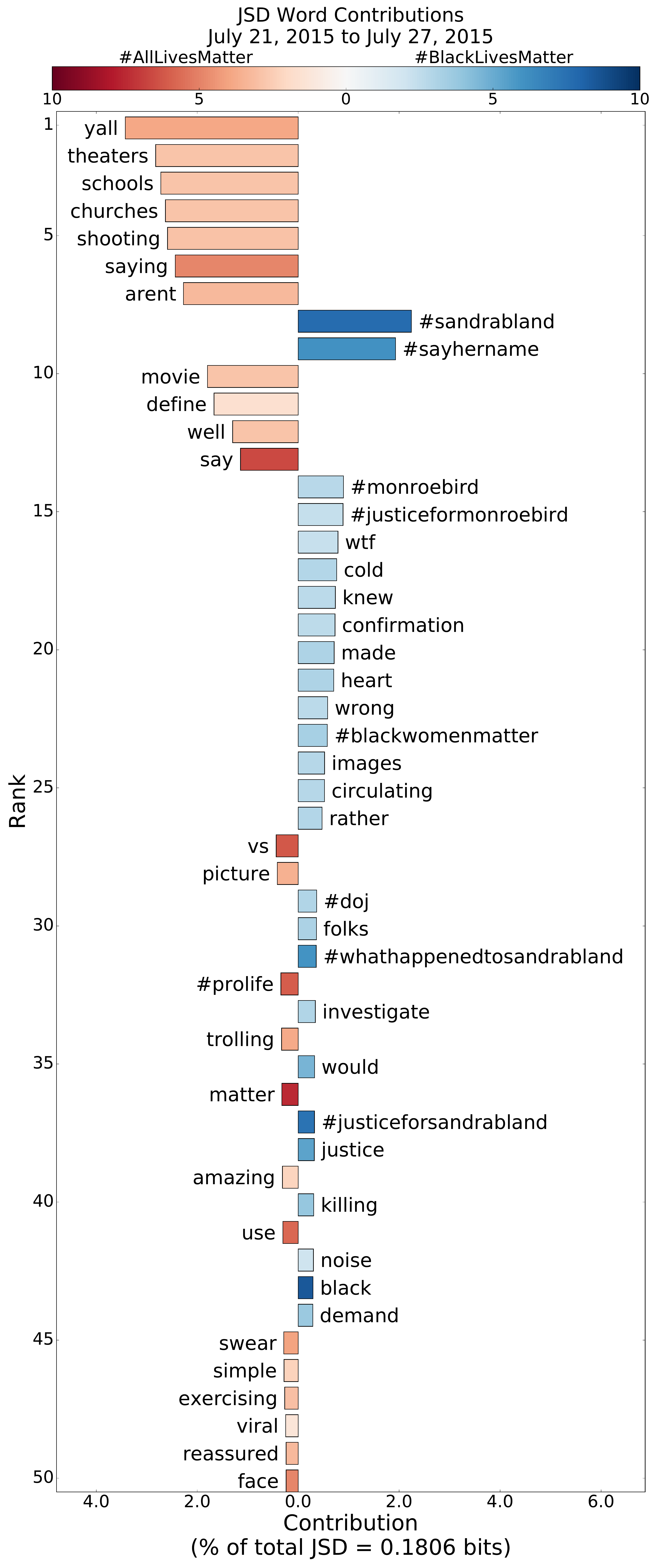}
\caption{Jensen-Shannon divergence word shift for the week encapsulating outrage over the death of Sandra Bland. See main text and the caption of  Fig.~\ref{2014-11-24-wordshift} for an explanation of the word shift graphs.}
\label{2015-07-21-wordshift}
\end{minipage}
\end{figure*}


\setcounter{figure}{0} 

\begin{figure*}[!]
\section{Hashtag Topic Networks}
\includegraphics[scale = .53]{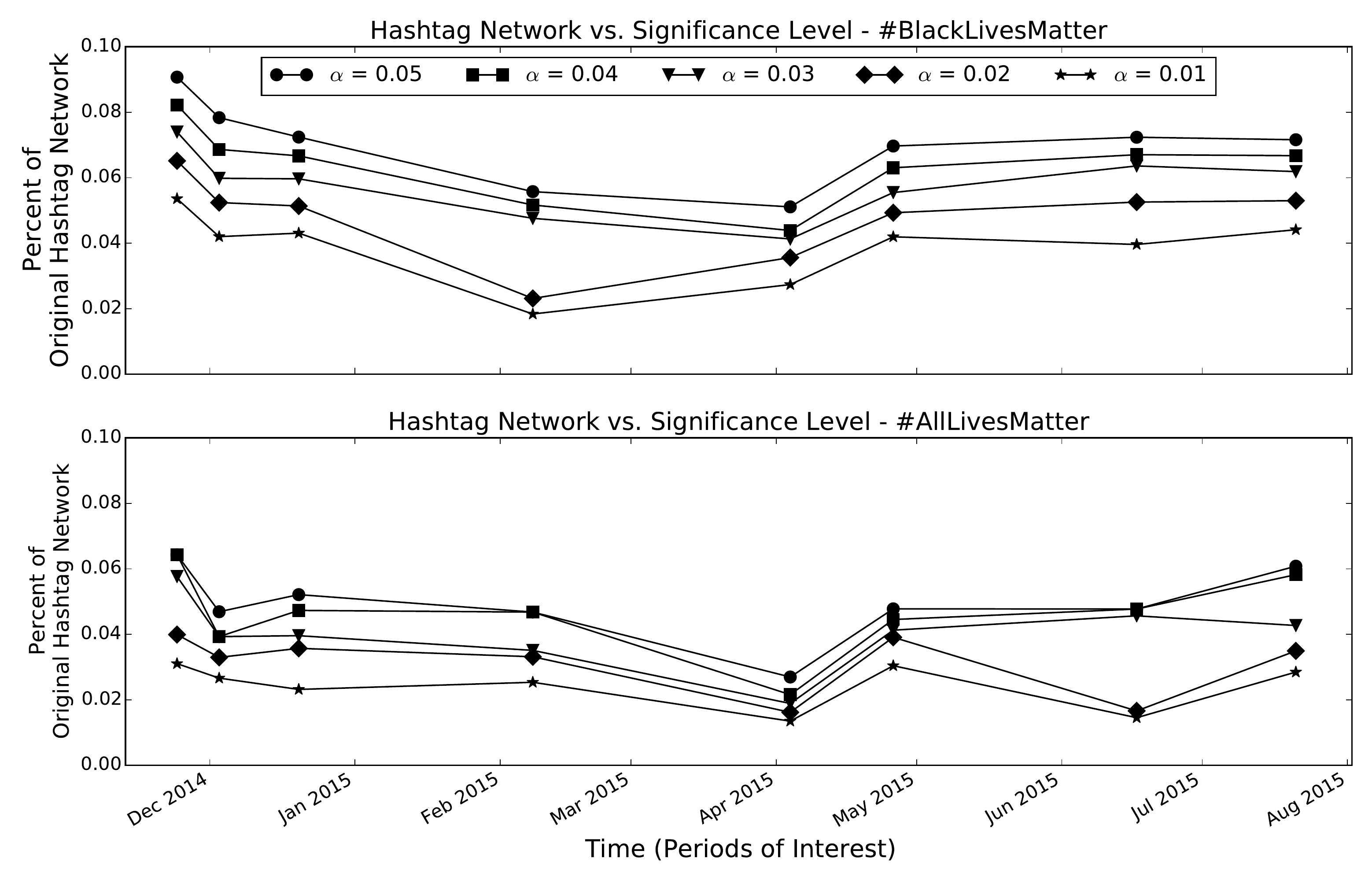}
\caption{Percent of original hashtag network maintained for \#BlackLivesMatter (top) and \#AllLivesMatter (bottom) at each of the periods of interest for varying levels of the disparity filter significance level. We wish to filter as much of the network as possible, while avoiding sudden reductions in the number of nodes in the network. Note, when going from $\alpha = 0.03$ to $\alpha = 0.02$, the February 8th \#BlackLivesMatter and July 21st \#AllLivesMatter networks fall in size by a factor of approximately one half. Therefore, we choose $\alpha = 0.03$.}
\label{significance-level}
\end{figure*}

\begin{table*}[!tbp]
\centering
\begin{tabular}{l|c|c|c|c}
\textbf{\#BlackLivesMatter}     & Nodes                 & \% Original Nodes     & Edges                 & Clustering           \\ \hline \hline
Nov. 24--Nov. 30, 2014            & 270                     & 8.21\%                     & 526                    & 0.0598               \\ \hline
Dec. 3--Dec. 9, 2014                & 389                     & 6.86\%                      & 912                    & 0.0645               \\ \hline
Dec. 20--Dec. 26, 2014            & 209                     & 6.66\%                     & 439                     & 0.1537               \\ \hline
Feb. 8--Feb. 14, 2015              & 76                      & 5.16\%                     & 107                       & 0.1744               \\ \hline
Apr. 4--Apr. 10, 2015               & 85                       & 4.38\%                     & 121                    & 0.0893               \\ \hline
Apr. 26--May 2, 2015               & 266                     & 6.30\%                     & 547                    & 0.1030               \\ \hline
Jun. 17--Jun. 23, 2015             & 176                    & 6.70\%                     & 269                    & 0.07690               \\ \hline
Jul. 21--Jul. 27, 2015               & 233                     & 6.67\%                     & 435                    & 0.0926               \\ \hline
\textbf{\#AllLivesMatter }      & \multicolumn{1}{l|}{} & \multicolumn{1}{l|}{} & \multicolumn{1}{l|}{} & \multicolumn{1}{l}{} \\ \hline \hline
Nov. 24--Nov. 30, 2014          & 29                       & 6.43\%                     & 40                      & 0.1200               \\ \hline
Dec. 3--Dec. 9, 2014               & 31                       & 3.92\%                     & 51                      & 0.2801               \\ \hline
Dec. 20--Dec. 26, 2014           & 49                       & 4.72\%                     & 86                      & 0.2626               \\ \hline
Feb. 8--Feb. 14, 2015            & 24                       & 4.67\%                     & 30                       & 0.1521               \\ \hline
Apr. 4--Apr. 10, 2015              & 8                        & 2.15\%                     & 7                        & 0.0000               \\ \hline
Apr. 26--May 2, 2015             & 41                       & 4.45\%                     & 71                      & 0.1967               \\ \hline
Jun. 17--Jun. 23, 2015           & 23                       & 4.77\%                     & 33                      & 0.4909               \\ \hline
Jul. 21--Jul. 27, 2015              & 45                       & 5.82\%                     & 61                      & 0.1010              
\end{tabular}
\caption{Summary statistics for topic networks created from the full hashtag networks using the disparity filter at the significance level $\protect\alpha$ = 0.04 \cite{serrano2009extracting}.}
\label{backbone-stats-alpha4}
\end{table*}

\begin{table*}[!tbp]
\centering
\begin{tabular}{l|c|c|c|c}
\textbf{\#BlackLivesMatter}     & Nodes                 & \% Original Nodes     & Edges                 & Clustering           \\ \hline \hline
Nov. 24--Nov. 30, 2014            & 298                     & 9.07\%                     & 583                    & 0.0590               \\ \hline
Dec. 3--Dec. 9, 2014                & 444                     & 7.83\%                      & 1025                    & 0.0595               \\ \hline
Dec. 20--Dec. 26, 2014            & 227                     & 7.24\%                     & 485                     & 0.1494               \\ \hline
Feb. 8--Feb. 14, 2015              & 82                       & 5.57\%                     & 119                       & 0.1946               \\ \hline
Apr. 4--Apr. 10, 2015               & 99                       & 5.10\%                     & 139                    & 0.0905               \\ \hline
Apr. 26--May 2, 2015               & 294                     & 6.96\%                     & 607                    & 0.0976               \\ \hline
Jun. 17--Jun. 23, 2015             & 190                    & 7.23\%                     & 305                    & 0.0795               \\ \hline
Jul. 21--Jul. 27, 2015               & 250                     & 7.15\%                     & 475                   & 0.0898               \\ \hline
\textbf{\#AllLivesMatter }      & \multicolumn{1}{l|}{} & \multicolumn{1}{l|}{} & \multicolumn{1}{l|}{} & \multicolumn{1}{l}{} \\ \hline \hline
Nov. 24--Nov. 30, 2014          & 29                       & 6.43\%                     & 40                      & 0.1200               \\ \hline
Dec. 3--Dec. 9, 2014               & 37                       & 4.68\%                     & 62                      & 0.2456               \\ \hline
Dec. 20--Dec. 26, 2014           & 54                       & 5.21\%                     & 94                      & 0.2500               \\ \hline
Feb. 8--Feb. 14, 2015            & 24                       & 4.67\%                     & 30                       & 0.1521               \\ \hline
Apr. 4--Apr. 10, 2015              & 10                       & 2.69\%                     & 9                        & 0.0000               \\ \hline
Apr. 26--May 2, 2015             & 44                       & 4.77\%                     & 76                      & 0.1944               \\ \hline
Jun. 17--Jun. 23, 2015           & 23                       & 4.77\%                     & 33                      & 0.4909               \\ \hline
Jul. 21--Jul. 27, 2015              & 47                       & 6.08\%                     & 66                      & 0.1151              
\end{tabular}
\caption{Summary statistics for topic networks created from the full hashtag networks using the disparity filter at the significance level $\protect\alpha$ = 0.05 \cite{serrano2009extracting}.}
\label{backbone-stats-alpha5}
\end{table*}

\begin{table*}[!htb]
\scriptsize
\centering
\begin{tabular}{ll|l|c|l|l|l|c|l}
                                           & \multicolumn{4}{c|}{\textbf{\#BlackLivesMatter}}                                                                                                                                                                                                                                                                                                                                                               & \multicolumn{4}{c}{\textbf{\#AllLivesMatter}}                                                                                                                                                                                                                                                                                                                                                                                        \\
\multicolumn{1}{l|}{}                      & \multicolumn{2}{c|}{Top Hashtags}                                                                                                                                                                                                             & \multicolumn{2}{c|}{Betweenness}                                                                                                                      & \multicolumn{2}{c|}{Top Hashtags}                                                                                                                                                                                                                                     & \multicolumn{2}{c}{Betweenness}                                                                                                                     \\ \hline \hline
\multicolumn{1}{l|}{Nov. 24--Nov. 30, 2014} & \multicolumn{2}{l|}{\begin{tabular}[c]{@{}l@{}}1. ferguson\\ 2. mikebrown\\ 3. standup\\ 4. tamirrice\\ 5. truth\\ 6. fergusondecision\\ 7. arrestdarrenwilson\\ 8. justiceformikebrown\\ 9. shutitdownatl\\ 10. alllivesmatter\end{tabular}} & \multicolumn{2}{c|}{\begin{tabular}[c]{@{}c@{}}0.8356\\ 0.1584\\ 0.0845\\ 0.0835\\ 0.0804\\ 0.7935\\ 0.0716\\ .0.0704\\ 0.0645\\ 0.0488\end{tabular}} & \multicolumn{2}{l|}{\begin{tabular}[c]{@{}l@{}}1. blacklivesmatter\\ 2. ferguson\\ 3. policebrutality\\ 4. mikebrown\\ 5. fergusondecision\end{tabular}}                                                                                                              & \multicolumn{2}{c}{\begin{tabular}[c]{@{}c@{}}0.6116\\ 0.4550\\ 0.1733\\ 0.1566\\ 0.1566\end{tabular}}                                              \\ \hline
\multicolumn{1}{l|}{Dec. 3--Dec. 9, 2014}   & \multicolumn{2}{l|}{\begin{tabular}[c]{@{}l@{}}1. icantbreathe\\ 2. ericgarner\\ 3. ferguson\\ 4. civilrights\\ 5. shutitdown\\ 6. lebronjames\\ 7. rip\\ 8. cantbreathe\\ 9. mikebrown\\ 10. nojusticenopeace\end{tabular}}                  & \multicolumn{2}{c|}{\begin{tabular}[c]{@{}c@{}}0.4434\\ 0.4126\\ 0.2706\\ 0.0952\\ 0.0939\\ 0.0823\\ 0.0654\\ 0.0624\\ 0.0611\\ 0.0514\end{tabular}}  & \multicolumn{2}{l|}{\begin{tabular}[c]{@{}l@{}}1. blacklivesmatter\\ 2. shutitdown\\ 3. ferguson\\ 4. ericgarner\\ 5. policebrutality\\ 6. tcot\\ 7. icantbreathe\\ 8. miami\\ 9. crimingwhilewhite\\ 10. handsupdontshoot\end{tabular}}                              & \multicolumn{2}{c}{\begin{tabular}[c]{@{}c@{}}0.5153\\ 0.3490\\ 0.2827\\ 0.2750\\ 0.2045\\ 0.1908\\ 0.1693\\ 0.0667\\ 0.0667\\ 0.0114\end{tabular}} \\ \hline
\multicolumn{1}{l|}{Dec. 20--Dec. 26, 2014} & \multicolumn{2}{l|}{\begin{tabular}[c]{@{}l@{}}1. icantbreathe\\ 2. ferguson\\ 3. antoniomartin\\ 4. shutitdown\\ 5. ny\\ 6. handsupdontshoot\\ 7. racism\\ 8. obama\\ 9. mlk\\ 10. justice\end{tabular}}                                     & \multicolumn{2}{c|}{\begin{tabular}[c]{@{}c@{}}0.5229\\ 0.2938\\ 0.2105\\ 0.1575\\ 0.1174\\ 0.1012\\ 0.0948\\ 0.9411\\ 0.0702\\ 0.0640\end{tabular}}  & \multicolumn{2}{l|}{\begin{tabular}[c]{@{}l@{}}1. nypd\\ 2. policelivesmatter\\ 3. blacklivesmatter\\ 4. icantbreathe\\ 5. bluelivesmatter\\ 6. ferguson\\ 7. handsupdontshoot\\ 8. nyc\\ 9. nojusticenopeace\\ 10. ericgarner\end{tabular}}                          & \multicolumn{2}{c}{\begin{tabular}[c]{@{}c@{}}0.4884\\ 0.3307\\ 0.3064\\ 0.2461\\ 0.2012\\ 0.1833\\ 0.1679\\ 0.1461\\ 0.0987\\ 0.0910\end{tabular}} \\ \hline
\multicolumn{1}{l|}{Feb. 8--Feb. 14, 2015}  & \multicolumn{2}{l|}{\begin{tabular}[c]{@{}l@{}}1. blackhistorymonth\\ 2. grammys\\ 3. bhm\\ 4. muslimlivesmatter\\ 5. alllivesmatter\\ 6. bluelivesmatter\\ 7. ferguson\\ 8. icantbreathe\\ 9. ericgarner\\ 10. racist\end{tabular}}          & \multicolumn{2}{c|}{\begin{tabular}[c]{@{}c@{}}0.6504\\ 0.5549\\ 0.4987\\ 0.4961\\ 0.4769\\ 0.3938\\ 0.3226\\ 0.2779\\ 0.1841\\ 0.1445\end{tabular}}  & \multicolumn{2}{l|}{\begin{tabular}[c]{@{}l@{}}1. muslimlivesmatter\\ 2. blacklivesmatter\\ 3. chapelhillshooting\\ 4. jewishlivesmatter\\ 5. whitelivesmatter\\ 6. ourthreewinners\end{tabular}}                                                                     & \multicolumn{2}{c}{\begin{tabular}[c]{@{}c@{}}0.5661\\ 0.3750\\ 0.3161\\ 0.2279\\ 0.01764\\ 0.1323\end{tabular}}                                    \\ \hline
\multicolumn{1}{l|}{Apr. 4--Apr. 10, 2015}  & \multicolumn{2}{l|}{\begin{tabular}[c]{@{}l@{}}1. walterscott\\ 2. blacktwitter\\ 3. icantbreathe\\ 4. alllivesmatter\\ 5. ripwalterscott\\ 6. p2\\ 7. tcot\\ 8. kendrickjohnson\\ 9. ferguson\\ 10. racism\end{tabular}}                     & \multicolumn{2}{c|}{\begin{tabular}[c]{@{}c@{}}0.8083\\ 0.2307\\ 0.1502\\ 0.1488\\ 0.1467\\ 0.1061\\ 0.0645\\ 0.0645\\ 0.0592\\ 0.0558\end{tabular}}  & \multicolumn{2}{l|}{1. blacklivesmatter}                                                                                                                                                                                                                              & \multicolumn{2}{c}{1.0000}                                                                                                                          \\ \hline
\multicolumn{1}{l|}{Apr. 26--May 2, 2015}   & \multicolumn{2}{l|}{\begin{tabular}[c]{@{}l@{}}1. freddiegray\\ 2. baltimore\\ 3. blackpower\\ 4. baltimoreriots\\ 5. baltimoreuprising\\ 6. ferguson\\ 7. solidarity\\ 8. baltimore\\ 9. mayday\\ 10. alllivesmatter\end{tabular}}           & \multicolumn{2}{c|}{\begin{tabular}[c]{@{}c@{}}0.4534\\ 0.3441\\ 0.2258\\ 0.2120\\ 0.1801\\ 0.1518\\ 0.1159\\ 0.1002\\ 0.0842\\ 0.0833\end{tabular}}  & \multicolumn{2}{l|}{\begin{tabular}[c]{@{}l@{}}1. blacklivesmatter\\ 2. peace\\ 3. baltimoreriots\\ 4. baltimore\\ 5. justice\\ 6. policelivesmatter\\ 7. freddiegray\\ 8. bluelivesmatter\\ 9. baltimoreuprising\\ 10. asianlivesmatter\end{tabular}}                & \multicolumn{2}{c}{\begin{tabular}[c]{@{}c@{}}0.6531\\ 0.3198\\ 0.2515\\ 0.2289\\ 0.1321\\ 0.1118\\ 0.0975\\ 0.0720\\ 0.0540\\ 0.0270\end{tabular}} \\ \hline
\multicolumn{1}{l|}{Jun. 17--Jun. 23, 2015} & \multicolumn{2}{l|}{\begin{tabular}[c]{@{}l@{}}1. charlestonshooting\\ 2. charleston\\ 3. tcot\\ 4. confederateflag\\ 5. racheldolezal\\ 6. blacktwitter\\ 7. usa\\ 8. baltimore\\ 9. love\\ 10. staywoke\end{tabular}}                       & \multicolumn{2}{c|}{\begin{tabular}[c]{@{}c@{}}0.6321\\ 0.4336\\ 0.1727\\ 0.1547\\ 0.1168\\ 0.1047\\ 0.0943\\ 0.0904\\ 0.0794\\ 0.0792\end{tabular}}  & \multicolumn{2}{l|}{\begin{tabular}[c]{@{}l@{}}1. charlestonshooting\\ 2. blacklivesmatter\\ 3. bluelivesmatter\\ 4. gunsense\\ 5. 2a\\ 6. pjnet\\ 7. gohomederay\\ 8. ferguson\\ 9. wakeupamerica\\ 10. tcot\end{tabular}}                                           & \multicolumn{2}{c}{\begin{tabular}[c]{@{}c@{}}0.6667\\ 0.6238\\ 0.4968\\ 0.2571\\ 0.1857\\ 0.1809\\ 0.0952\\ 0.0952\\ 0.0015\\ 0.0015\end{tabular}} \\ \hline
\multicolumn{1}{l|}{Jul. 21--Jul. 27, 2015} & \multicolumn{2}{l|}{\begin{tabular}[c]{@{}l@{}}1. sandrabland\\ 2. sayhername\\ 3. justiceforsandrabland\\ 4. blacktwitter\\ 5. alllivesmatter\\ 6. wewantanswers\\ 7. blackpower\\ 8. sandra\\ 9. m4bl\\ 10. blm\end{tabular}}               & \multicolumn{2}{c|}{\begin{tabular}[c]{@{}c@{}}0.7038\\ 0.1985\\ 0.1533\\ 0.1385\\ 0.1154\\ 0.1153\\ 0.1128\\ 0.0891\\ 0.0731\\ 0.0656\end{tabular}}  & \multicolumn{2}{l|}{\begin{tabular}[c]{@{}l@{}}1. blacklivesmatter\\ 2. tcot\\ 3. policebrutality\\ 4. uniteblue\\ 5. sandrabland\\ 6. pjnet\\ 7. justiceforsandrabland\\ 8. defundplannedparenthood\\ 9. wakeupamerica\\ 10. whathappenedtosandrabland\end{tabular}} & \multicolumn{2}{c}{\begin{tabular}[c]{@{}c@{}}0.7842\\ 0.3689\\ 0.2177\\ 0.1794\\ 0.1754\\ 0.1411\\ 0.1209\\ 0.1189\\ 0.1028\\ 0.0625\end{tabular}}
\end{tabular}
\caption{The top 10 hashtags in the topic networks as determined by betweenness centrality for each time period. Some \#AllLivesMatter topic networks have less than 10 top nodes due to the relatively small size of the networks. For example, in the period of April 4th, the \#AllLivesMatter network consists of \#blacklivesmatter as a hub with six edges connecting it to six other hashtags. Thus, \#blacklivesmatter is the only node visited when calculating paths for betweenness.}
\label{backbone-betweenness}
\end{table*}

\begin{table*}[!htb]
\scriptsize
\centering
\begin{tabular}{ll|c|l|c}
                                            & \multicolumn{2}{c|}{\textbf{\#BlackLivesMatter}}                                                                                                                                                                                                                                                                                                                     & \multicolumn{2}{c}{\textbf{\#AllLivesMatter}}                                                                                                                                                                                                                                                                                                                                                                    \\
\multicolumn{1}{l|}{}                       & \multicolumn{1}{c|}{Top Hashtags}                                                                                                                                                                                                  & PageRank                                                                                                                        & \multicolumn{1}{c|}{Top Hashtags}                                                                                                                                                                                                                                               & \multicolumn{1}{c}{PageRank}                                                                                                   \\ \hline \hline
\multicolumn{1}{l|}{Nov. 24--Nov. 30, 2014} & \begin{tabular}[c]{@{}l@{}}1. ferguson\\ 2. mikebrown\\ 3. fergusondecision\\ 4. shutitdown\\ 5. michaelbrown\\ 6. blackoutblackfriday\\ 7. boycottblackfriday\\ 8. justiceformikebrown\\ 9. notonedime\\ 10. america\end{tabular} & \begin{tabular}[c]{@{}c@{}}0.2299\\ 0.0778\\ 0.0504\\ 0.0343\\ 0.0258\\ 0.0238\\ 0.0203\\ 0.0197\\ 0.0176\\ 0.0161\end{tabular} & \begin{tabular}[c]{@{}l@{}}1. blacklivesmatter\\ 2. ferguson\\ 3. fergusondecision\\ 4. mikebrown\\ 5. nycprotest\\ 6. williamsburg\\ 7. brownlivesmatter\\ 8. whitelivesmatter\\ 9. sf\\ 10. blackfridayblackout\end{tabular}                                                  & \begin{tabular}[c]{@{}c@{}}0.2647\\ 0.2371\\ 0.0575\\ 0.0509\\ 0.0397\\ 0.0317\\ 0.0239\\ 0.0235\\ 0.0224\\ 0.0210\end{tabular} \\ \hline
\multicolumn{1}{l|}{Dec. 3--Dec. 9, 2014}   & \begin{tabular}[c]{@{}l@{}}1. ericgarner\\ 2. icantbreathe\\ 3. ferguson\\ 4. mikebrown\\ \\ 5. nypd\\ 6. shutitdown\\ 7. handsupdontshoot\\ 8. thisstopstoday\\ 9. seattle\\ 10. policebrutality\end{tabular}                     & \begin{tabular}[c]{@{}c@{}}0.1974\\ 0.1548\\ 0.0860\\ 0.0296\\ 0.0286\\ 0.0251\\ 0.0183\\ 0.0151\\ 0.0142\\ 0.0140\end{tabular} & \begin{tabular}[c]{@{}l@{}}1. blacklivesmatter\\ 2. icantbreathe\\ 3. ericgarner\\ 4. ferguson\\ 5. tcot\\ 6. wecantbreathe\\ 7. mikebrown\\ 8. handsupdontshoot\\ 9. rednationrising\\ 10. shutitdown\end{tabular}                                                             & \begin{tabular}[c]{@{}c@{}}0.2052\\ 0.1397\\ 0.1295\\ 0.0728\\ 0.0404\\ 0.0379\\ 0.0371\\ 0.0355\\ 0.0268\\ 0.0267\end{tabular} \\ \hline
\multicolumn{1}{l|}{Dec. 20--Dec. 26, 2014} & \begin{tabular}[c]{@{}l@{}}1. icantbreathe\\ 2. ferguson\\ 3. shutitdown\\ 4. antoniomartin\\ 5. ericgarner\\ 6. nypdlivesmatter\\ 7. alllivesmatter\\ 8. moa\\ 9. nypd\\ 10. mikebrown\end{tabular}                               & \begin{tabular}[c]{@{}c@{}}0.1043\\ 0.0533\\ 0.0450\\ 0.0397\\ 0.0333\\ 0.0315\\ 0.0256\\ 0.0253\\ 0.0247\\ 0.0227\end{tabular} & \begin{tabular}[c]{@{}l@{}}1. blacklivesmatter\\ 2. nypdlivesmatter\\ 3. nypd\\ 4. policelivesmatter\\ 5. ericgarner\\ 6. mikebrown\\ 7. bluelivesmatter\\ 8. icantbreathe\\ 9. nyc\\ 10. stolenlives\end{tabular}                                                              & \begin{tabular}[c]{@{}c@{}}0.2042\\ 0.0955\\ 0.0945\\ 0.0608\\ 0.0569\\ 0.0471\\ 0.0328\\ 0.0310\\ 0.0291\\ 0.0278\end{tabular} \\ \hline
\multicolumn{1}{l|}{Feb. 8--Feb. 14, 2015}  & \begin{tabular}[c]{@{}l@{}}1. muslimlivesmatter\\ 2. alllivesmatter\\ 3. handsupdontshoot\\ 4. grammys\\ 5. mikebrown\\ 6. blackhistorymonth\\ 7. beyhive\\ 8. ferguson\\ 9. blacktwitter\\ 10. chapelhillshooting\end{tabular}    & \begin{tabular}[c]{@{}c@{}}0.1418\\ 0.0730\\ 0.0686\\ 0.0575\\ 0.0485\\ 0.0467\\ 0.0447\\ 0.0338\\ 0.0273\\ 0.0244\end{tabular} & \begin{tabular}[c]{@{}l@{}}1. muslimlivesmatter\\ 2. blacklivesmatter\\ 3. chapelhillshooting\\ 4. butinacosmicsensenothingreallymatters\\ 5. jewishlivesmatter\\ 6. christianslivesmatter\\ 7. buddhistlivesmatter\\ 8. whitelivesmatter\\ 9. rip\\ 10. hatecrime\end{tabular} & \begin{tabular}[c]{@{}c@{}}0.3171\\ 0.1815\\ 0.1791\\ 0.0582\\ 0.0534\\ 0.0218\\ 0.0218\\ 0.0214\\ 0.0197\\ 0.0181\end{tabular} \\ \hline
\multicolumn{1}{l|}{Apr. 4--Apr. 10, 2015}  & \begin{tabular}[c]{@{}l@{}}1. walterscott\\ 2. ferguson\\ 3. ericgarner\\ 4. trayvonmartin\\ 5. blacktwitter\\ 6. icantbreathe\\ 7. mikebrown\\ 8. ftp\\ 9. alllivesmatter\\ 10. black\end{tabular}                                & \begin{tabular}[c]{@{}c@{}}0.2141\\ 0.0850\\ 0.0708\\ 0.0619\\ 0.0492\\ 0.0381\\ 0.0231\\ 0.0217\\ 0.0175\\ 0.0148\end{tabular} & \begin{tabular}[c]{@{}l@{}}1. blacklivesmatter\\ 2. walterscott\\ 3. muslimlivesmatter\\ 4. whitelivesmatter\\ 5. icantbreathe\\ 6. ripwalterscott\\ 7. ifnotnowwhen\end{tabular}                                                                                               & \begin{tabular}[c]{@{}c@{}}0.4710\\ 0.2253\\ 0.0705\\ 0.0667\\ 0.0629\\ 0.0516\\ 0.0516\end{tabular}                            \\ \hline
\multicolumn{1}{l|}{Apr. 26--May 2, 2015}   & \begin{tabular}[c]{@{}l@{}}1. freddiegray\\ 2. baltimore\\ 3. baltimoreuprising\\ 4. baltimoreriots\\ 5. alllivesmatter\\ 6. ericgarner\\ 7. ferguson\\ 8. mayday\\ 9. blackspring\\ 10. michaelbrown\end{tabular}                 & \begin{tabular}[c]{@{}c@{}}0.1362\\ 0.1154\\ 0.0850\\ 0.0787\\ 0.0153\\ 0.0132\\ 0.0128\\ 0.0115\\ 0.0108\\ 0.0094\end{tabular} & \begin{tabular}[c]{@{}l@{}}1. blacklivesmatter\\ 2. baltimoreriots\\ 3. baltimore\\ 4. freddiegray\\ 5. policelivesmatter\\ 6. baltimoreuprising\\ 7. tcot\\ 8. whitelivesmatter\\ 9. wakeupamerica\\ 10. prayforbaltimore\end{tabular}                                         & \begin{tabular}[c]{@{}c@{}}0.2104\\ 0.1416\\ 0.1115\\ 0.0799\\ 0.0427\\ 0.0419\\ 0.0327\\ 0.0304\\ 0.0205\\ 0.0192\end{tabular} \\ \hline
\multicolumn{1}{l|}{Jun. 17--Jun. 23, 2015} & \begin{tabular}[c]{@{}l@{}}1. charlestonshooting\\ 2. racism\\ 3. charleston\\ 4. whiteprivilege\\ 5. blacktwitter\\ 6. ferguson\\ 7. itsaracething\\ 8. usa\\ 9. southcarolina\\ 10. alllivesmatter\end{tabular}                  & \begin{tabular}[c]{@{}c@{}}0.1670\\ 0.0361\\ 0.0349\\ 0.0217\\ 0.0216\\ 0.0203\\ 0.0198\\ 0.0197\\ 0.0189\\ 0.0186\end{tabular} & \begin{tabular}[c]{@{}l@{}}1. blacklivesmatter\\ 2. iamame\\ 3. wakeupamerica\\ 4. pjnet\\ 5. bluelivesmatter\\ 6. tcot\\ 7. 2a\\ 8. charlestonshooting\\ 9. ohhillno\\ 10. cosproject\end{tabular}                                                                             & \begin{tabular}[c]{@{}c@{}}0.2059\\ 0.1423\\ 0.0699\\ 0.0656\\ 0.0648\\ 0.0634\\ 0.0602\\ 0.0392\\ 0.0378\\ 0.0378\end{tabular} \\ \hline
\multicolumn{1}{l|}{Jul. 21--Jul. 27, 2015} & \begin{tabular}[c]{@{}l@{}}1. sandrabland\\ 2. sayhername\\ 3. justiceforsandrabland\\ 4. blackwomenmatter\\ 5. doj\\ 6. blacktwitter\\ 7. unitedblue\\ 8. alllivesmatter\\ 9. whathappenedtosandrabland\\ 10. tcot\end{tabular}   & \begin{tabular}[c]{@{}c@{}}0.2013\\ 0.1449\\ 0.0422\\ 0.0352\\ 0.0238\\ 0.0201\\ 0.0169\\ 0.0163\\ 0.0159\\ 0.0132\end{tabular} & \begin{tabular}[c]{@{}l@{}}1. blacklivesmatter\\ 2. sandrabland\\ 3. pjnet\\ 4. defundpp\\ 5. justiceforsandrabland\\ 6. uniteblue\\ 7. sayhername\\ 8. defundplannedparenthood\\ 9. tcot\\ 10. prolife\end{tabular}                                                            & \begin{tabular}[c]{@{}c@{}}0.2302\\ 0.0996\\ 0.0661\\ 0.0526\\ 0.0505\\ 0.0486\\ 0.0472\\ 0.0470\\ 0.0456\\ 0.0344\end{tabular}
\end{tabular}
\caption{The top 10 hashtags in the topic networks as determined by PageRank for each time period. Some \#AllLivesMatter topic networks have less than 10 top nodes due to the relatively small size of the networks.}
\label{pagerank}
\end{table*}


\begin{figure*}
\centering
\includegraphics[scale=.7, trim = 50 0 0 0]{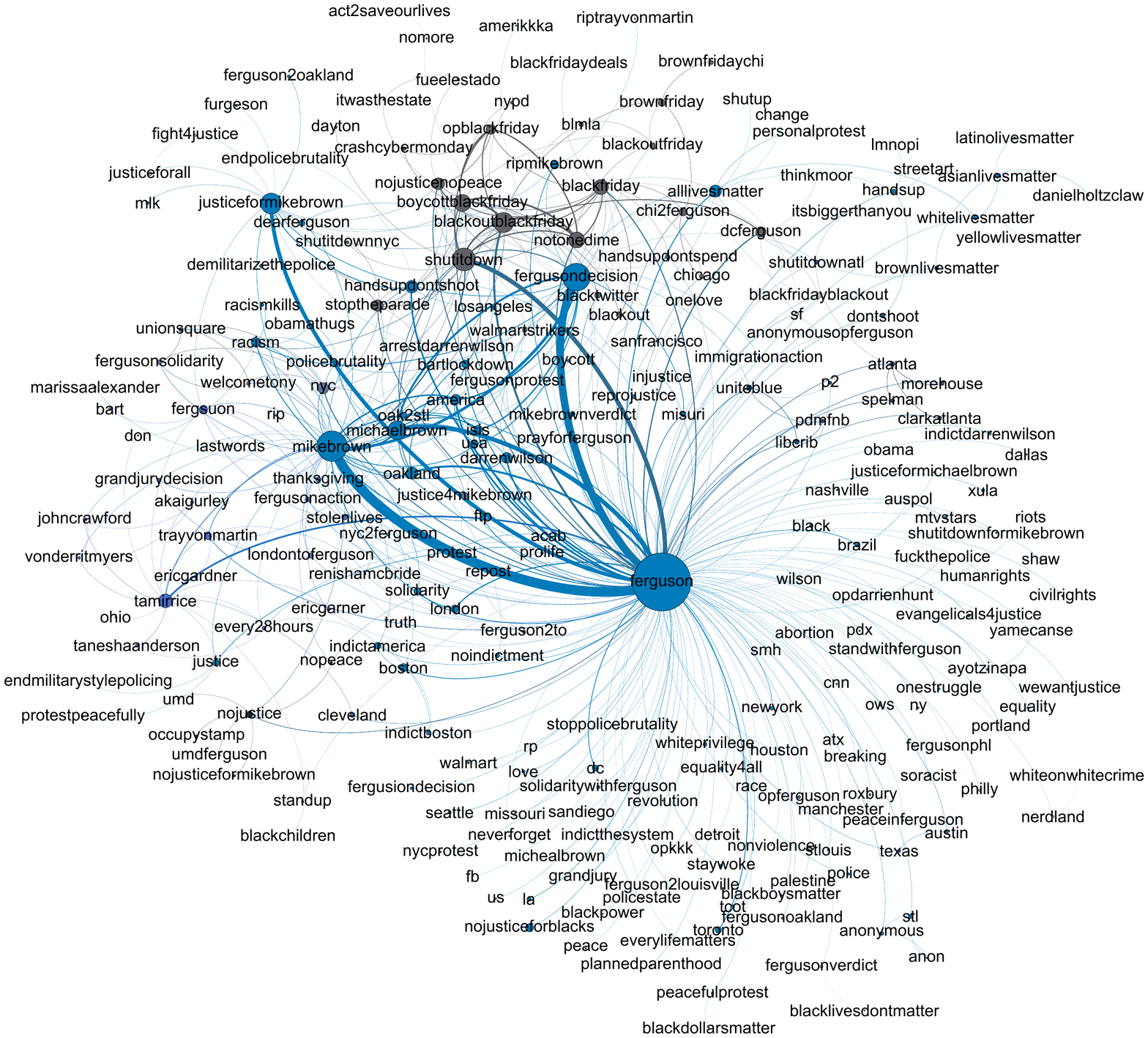}
\caption{\#BlackLivesMatter topic network for the week following the non-indictment of Darren Wilson.}
\label{2014-12-03-backbone-blacklives}
\end{figure*}

\begin{figure*}
\includegraphics[scale=.3]{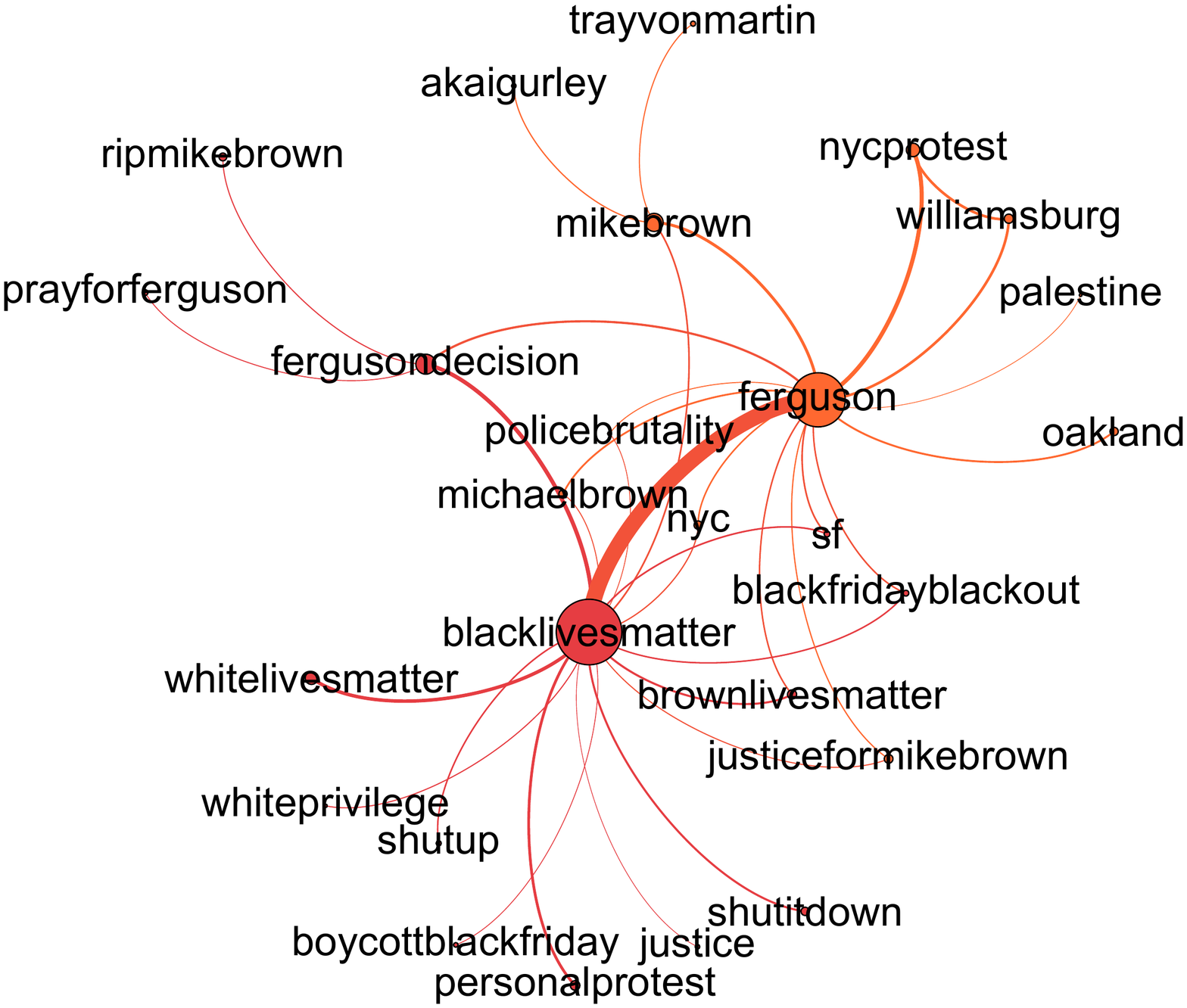}
\caption{\#AllLivesMatter topic network for the week following the non-indictment of Darren Wilson.}
\end{figure*}

\let\cleardoublepage\clearpage


\begin{figure*}
\centering
\includegraphics[scale=.7, trim = 50 0 0 0]{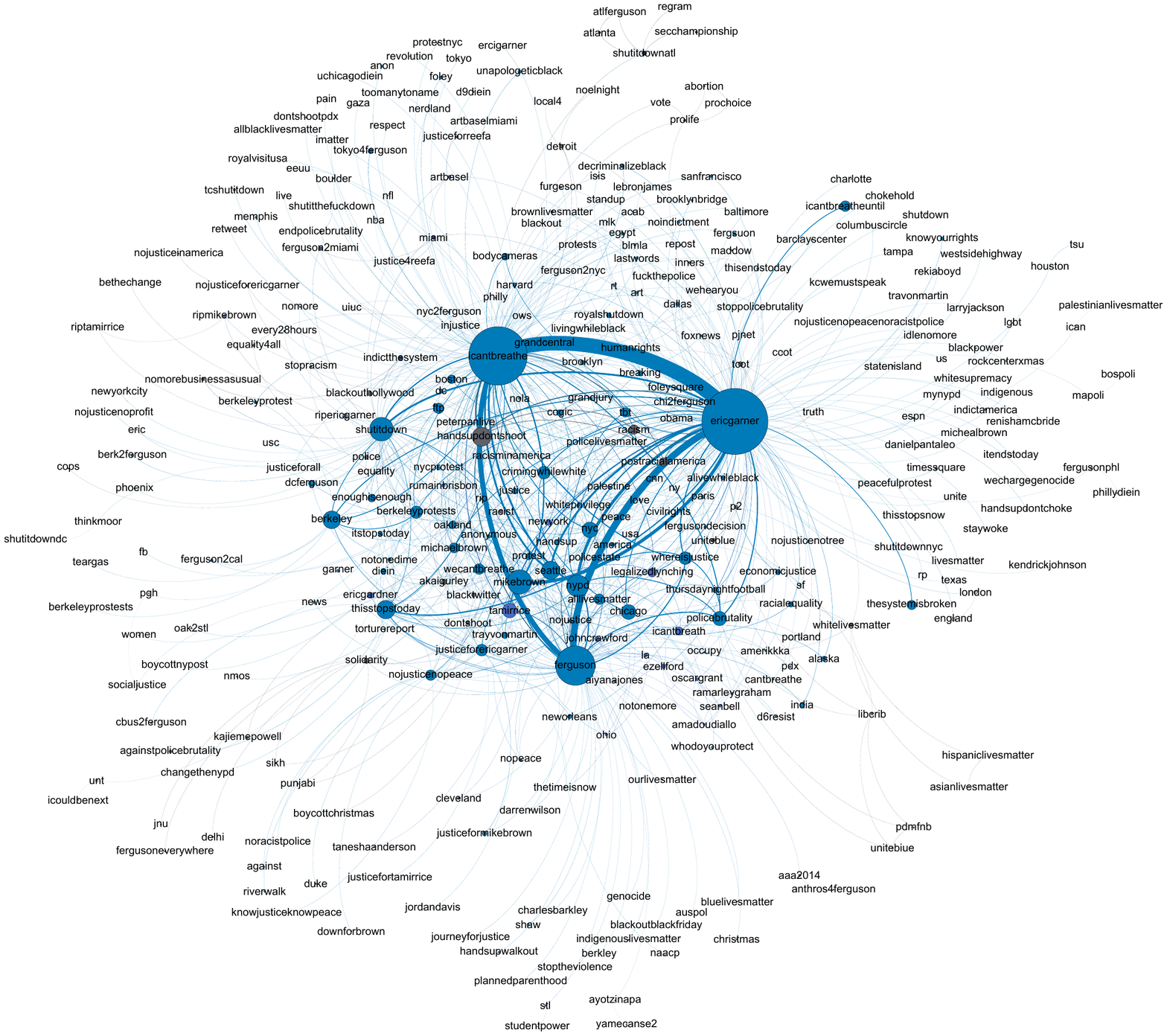}
\caption{\#BlackLivesMatter topic network for the week following the non-indictment of Daniel Pantaleo.}
\label{2014-12-03-backbone-blacklives}
\end{figure*}

\begin{figure*}
\includegraphics[scale=.3]{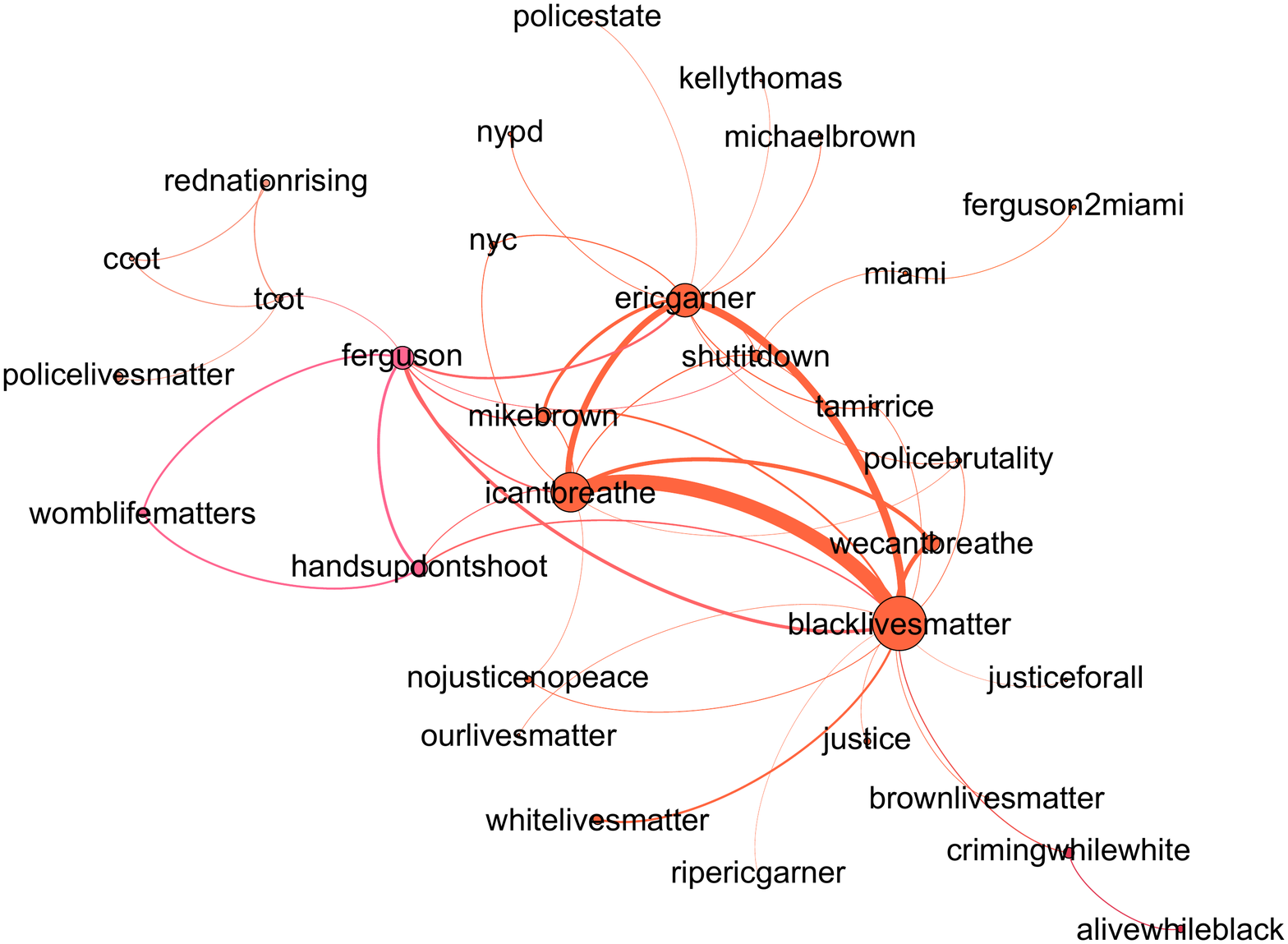}
\caption{\#AllLivesMatter topic network for the week following the non-indictment of Daniel Pantaleo.}
\end{figure*}

\let\cleardoublepage\clearpage 



\begin{figure*}
\centering
\includegraphics[scale=.65, trim = 20 0 0 0]{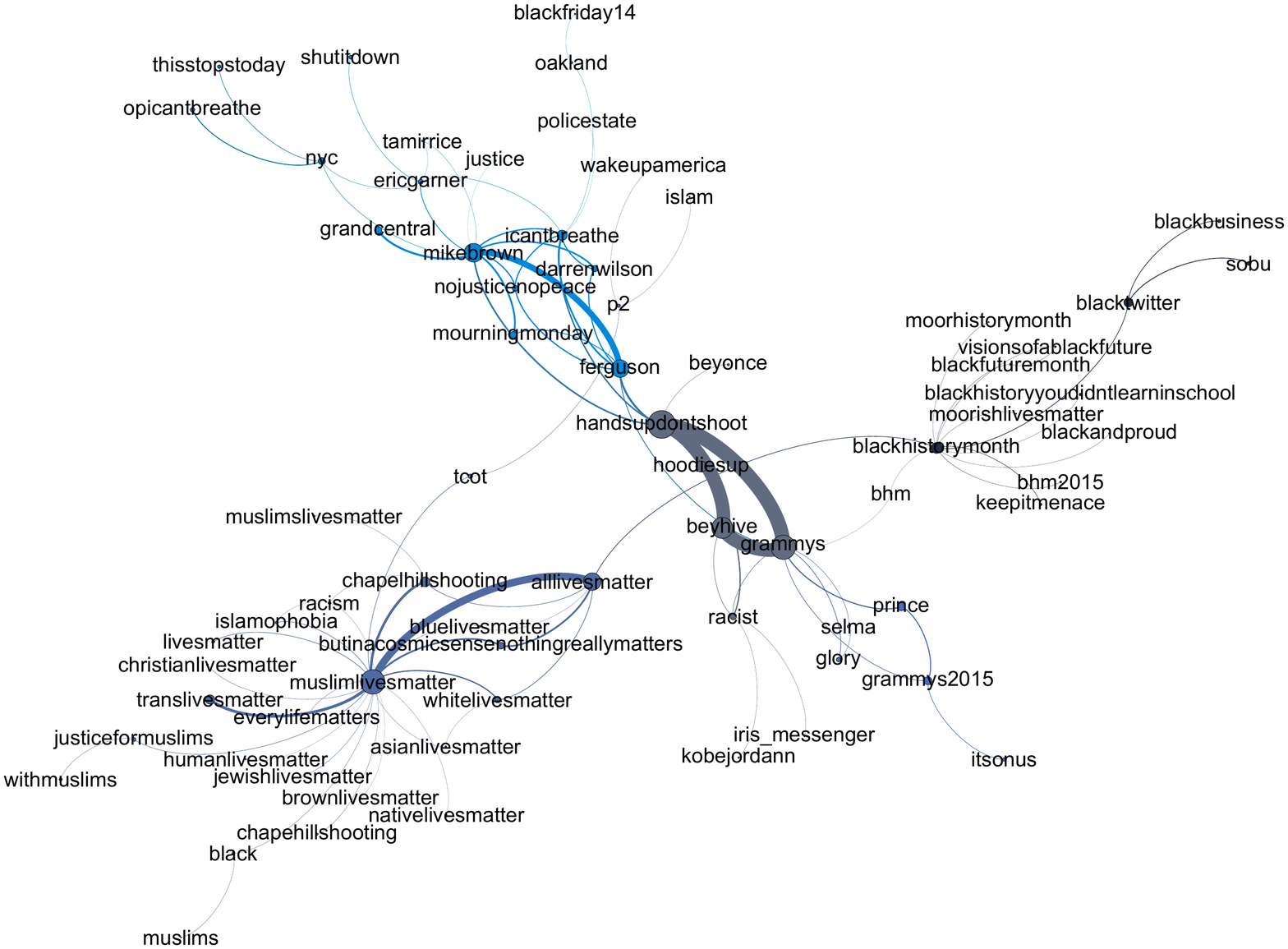}
\caption{\#BlackLivesMatter topic network for the week following the Chapel Hill shooting.}
\end{figure*}

\begin{figure*}
\includegraphics[scale=.35]{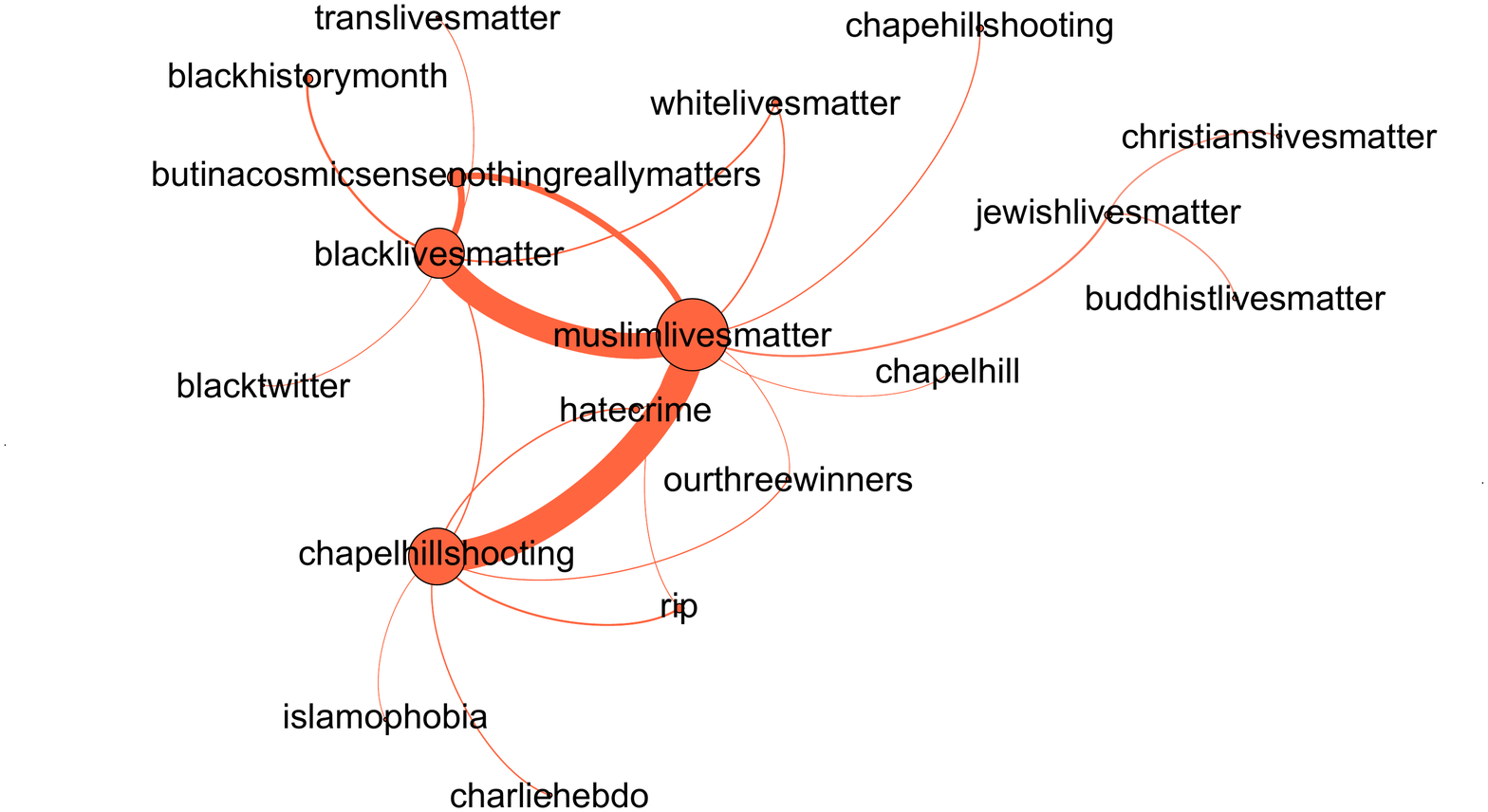}
\caption{\#AllLivesMatter topic network for the week following the Chapel Hill shooting.}
\label{2015-02-08-backbone-alllives}
\end{figure*}

\let\cleardoublepage\clearpage 


\begin{figure*}
\centering
\includegraphics[scale=.7, trim = 50 0 0 0]{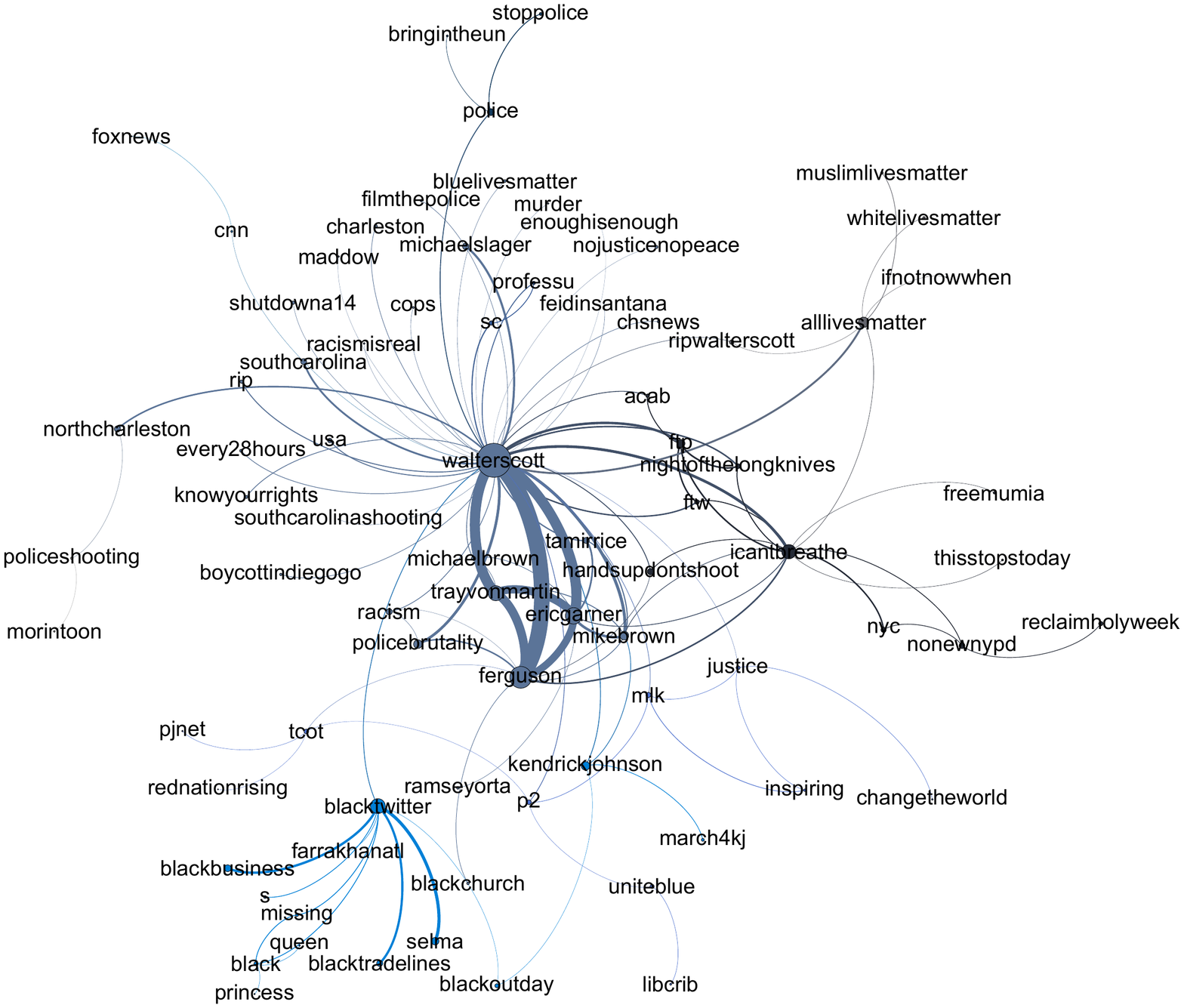}
\caption{\#BlackLivesMatter topic network for the week following the death of Walter Scott.}
\end{figure*}

\begin{figure*}
\includegraphics[scale=.3]{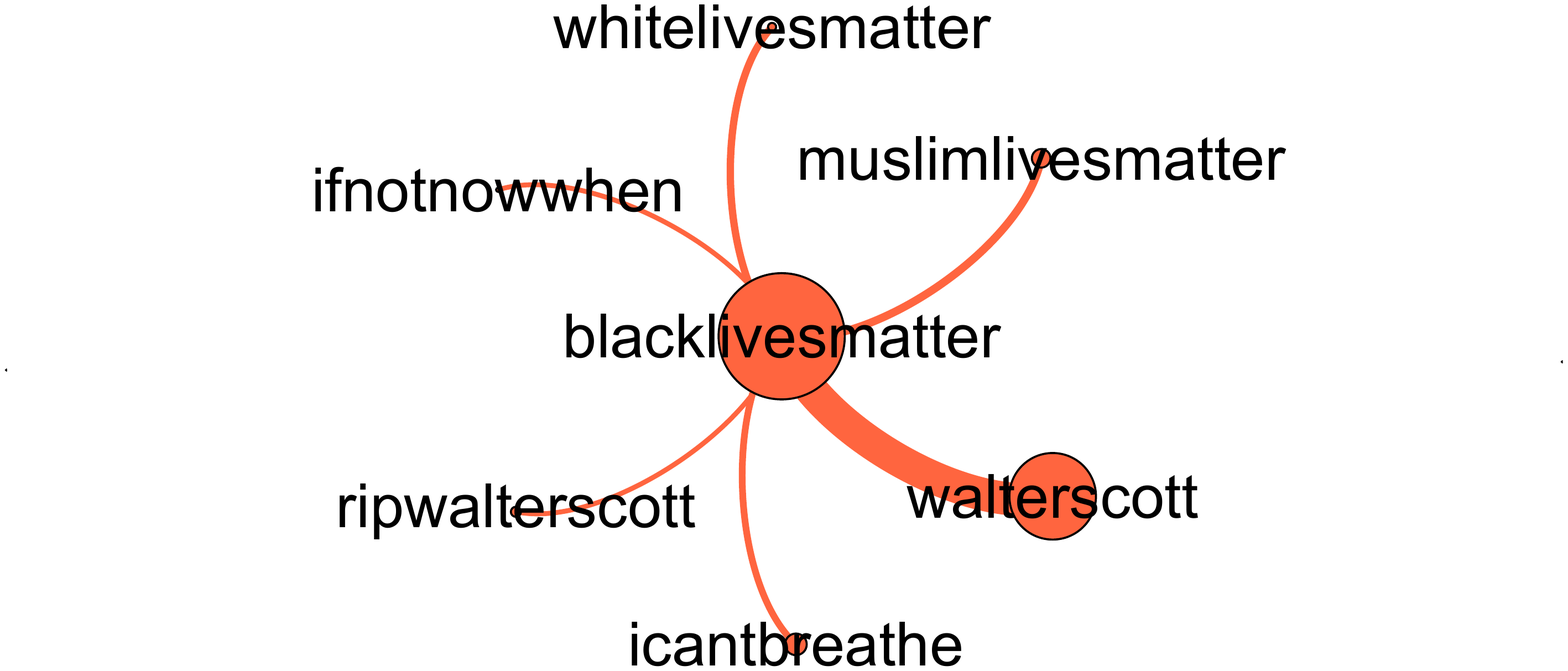}
\caption{\#AllLivesMatter topic network for the week following the death of Walter Scott.}
\end{figure*}

\let\cleardoublepage\clearpage 


\begin{figure*}
\centering
\includegraphics[scale=.65, trim = 50 0 0 0]{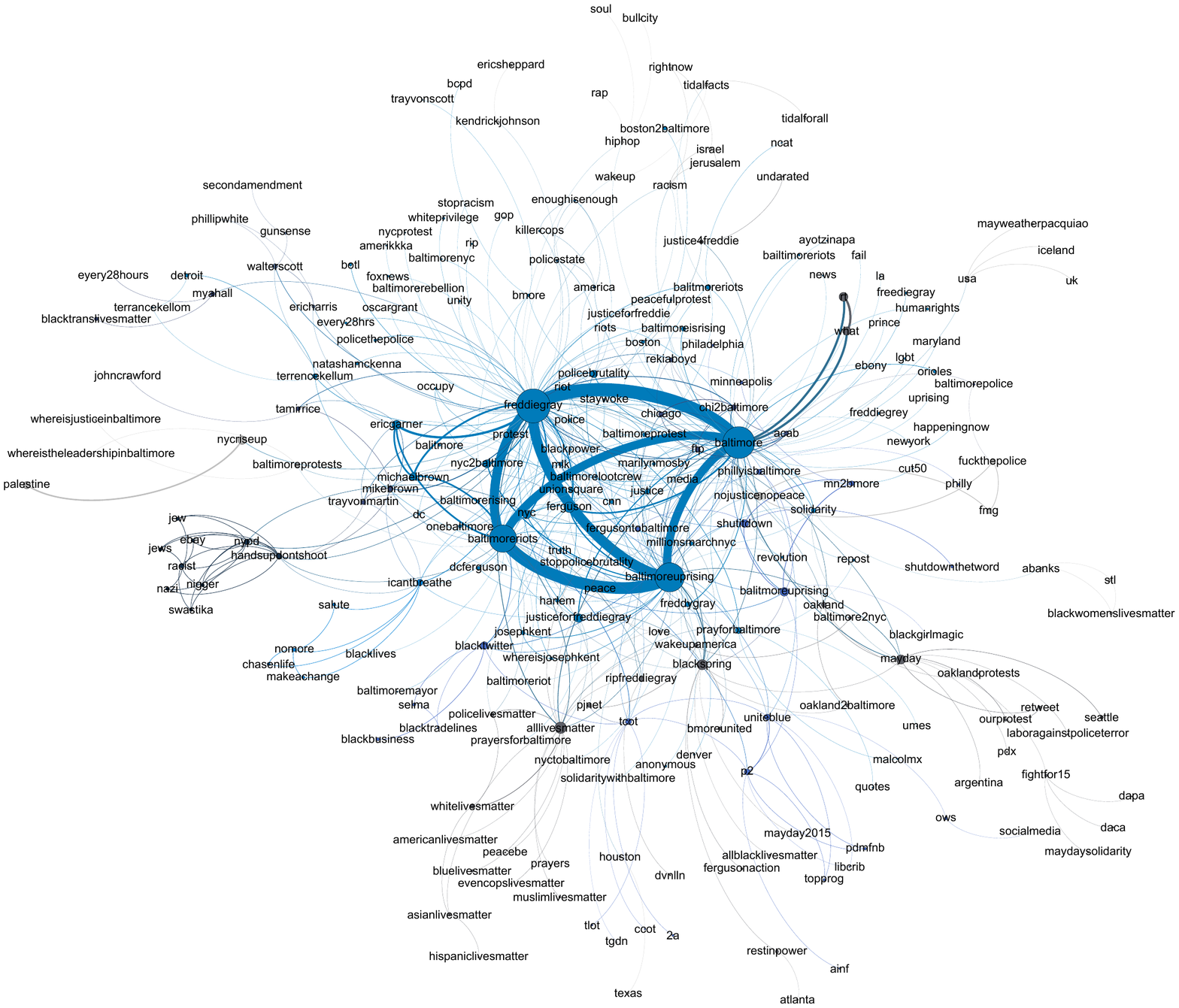}
\caption{\#BlackLivesMatter topic network for the week encapsulating the peak of the Baltimore protests.}
\end{figure*}

\begin{figure*}
\includegraphics[scale=.38]{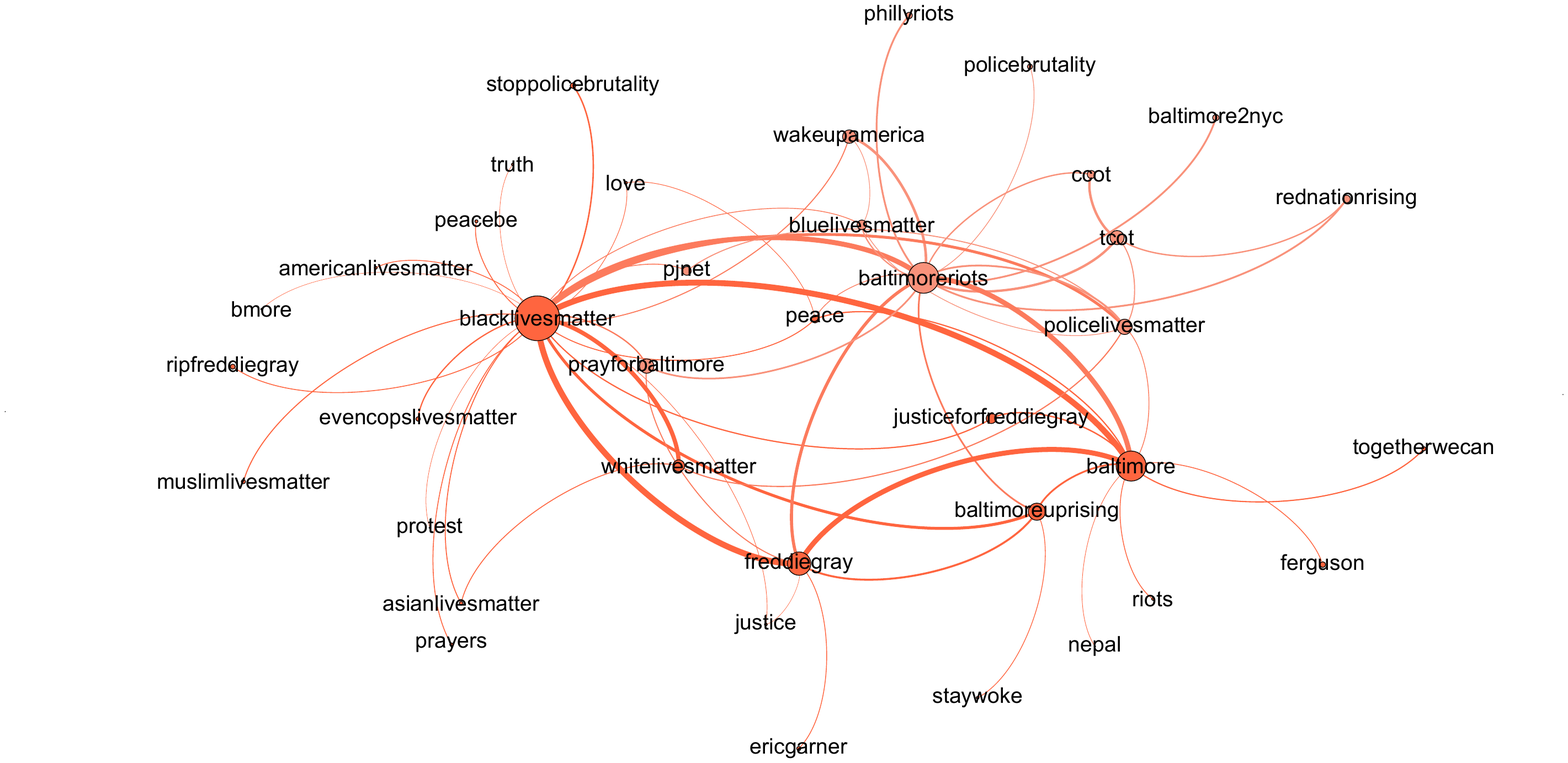}
\caption{\#AllLivesMatter topic network for the week encapsulating the peak of the Baltimore protests.}
\end{figure*}


\begin{figure*}
\centering
\includegraphics[scale=.65, trim = 10 0 0 0]{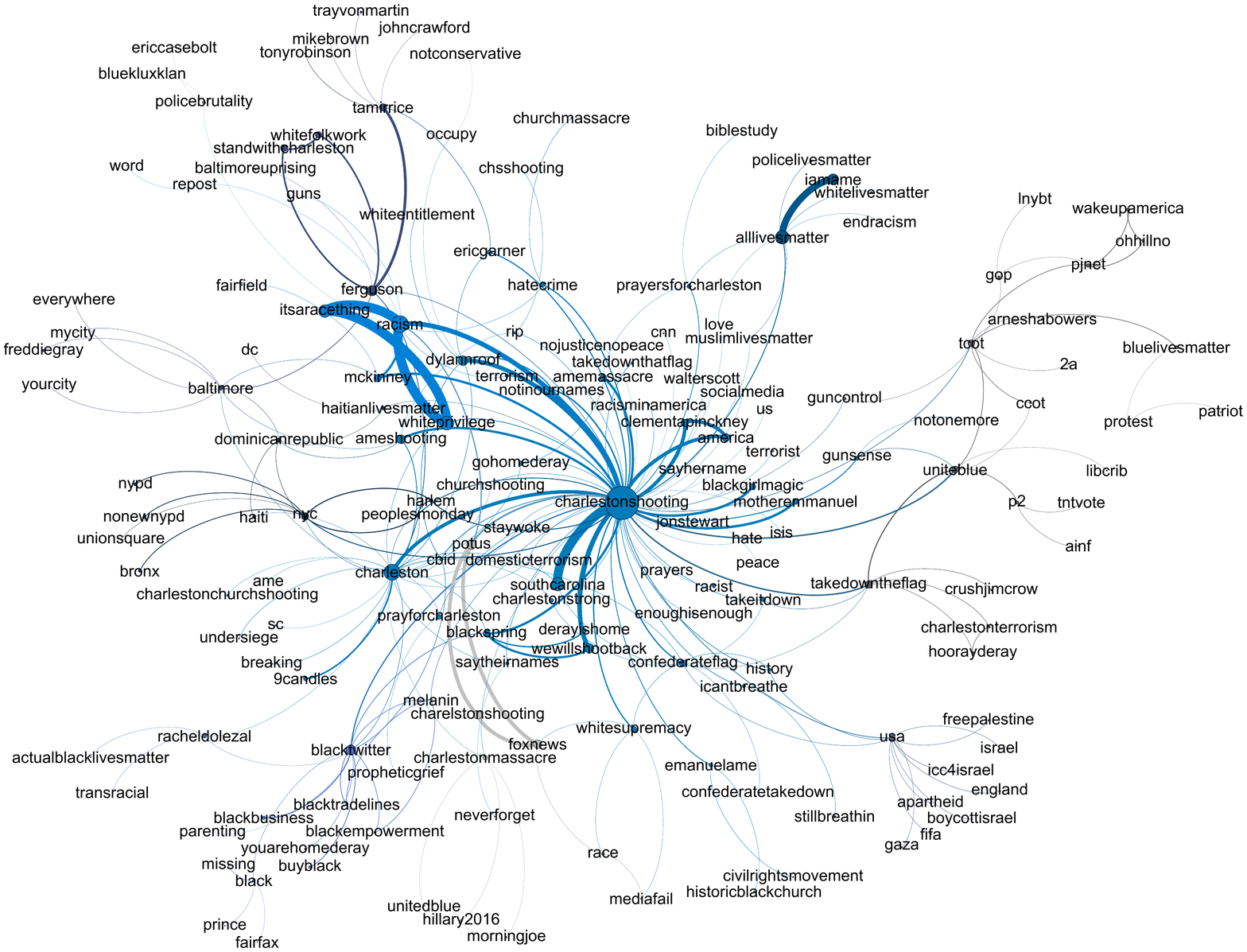}
\caption{\#BlackLivesMatter topic network for the week following the Charleston church shooting.}
\end{figure*}

\begin{figure*}
\includegraphics[scale=.38]{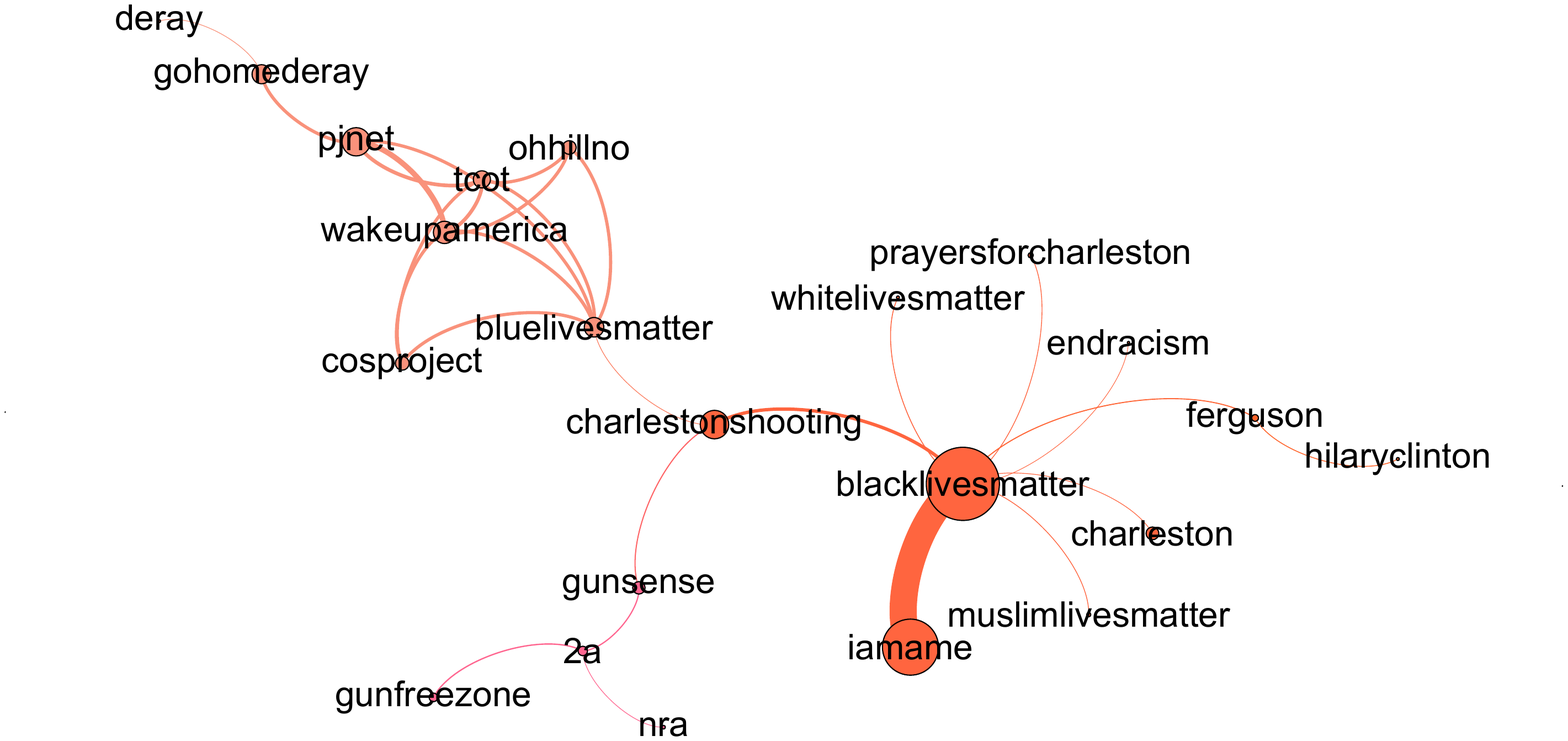}
\caption{\#AllLivesMatter topic network for the week following the Charleston church shoting.}
\label{2015-07-21-backbone-alllives}
\end{figure*}

\clearpage


\begin{figure*}
\centering
\includegraphics[scale=.65, trim = 10 0 0 0]{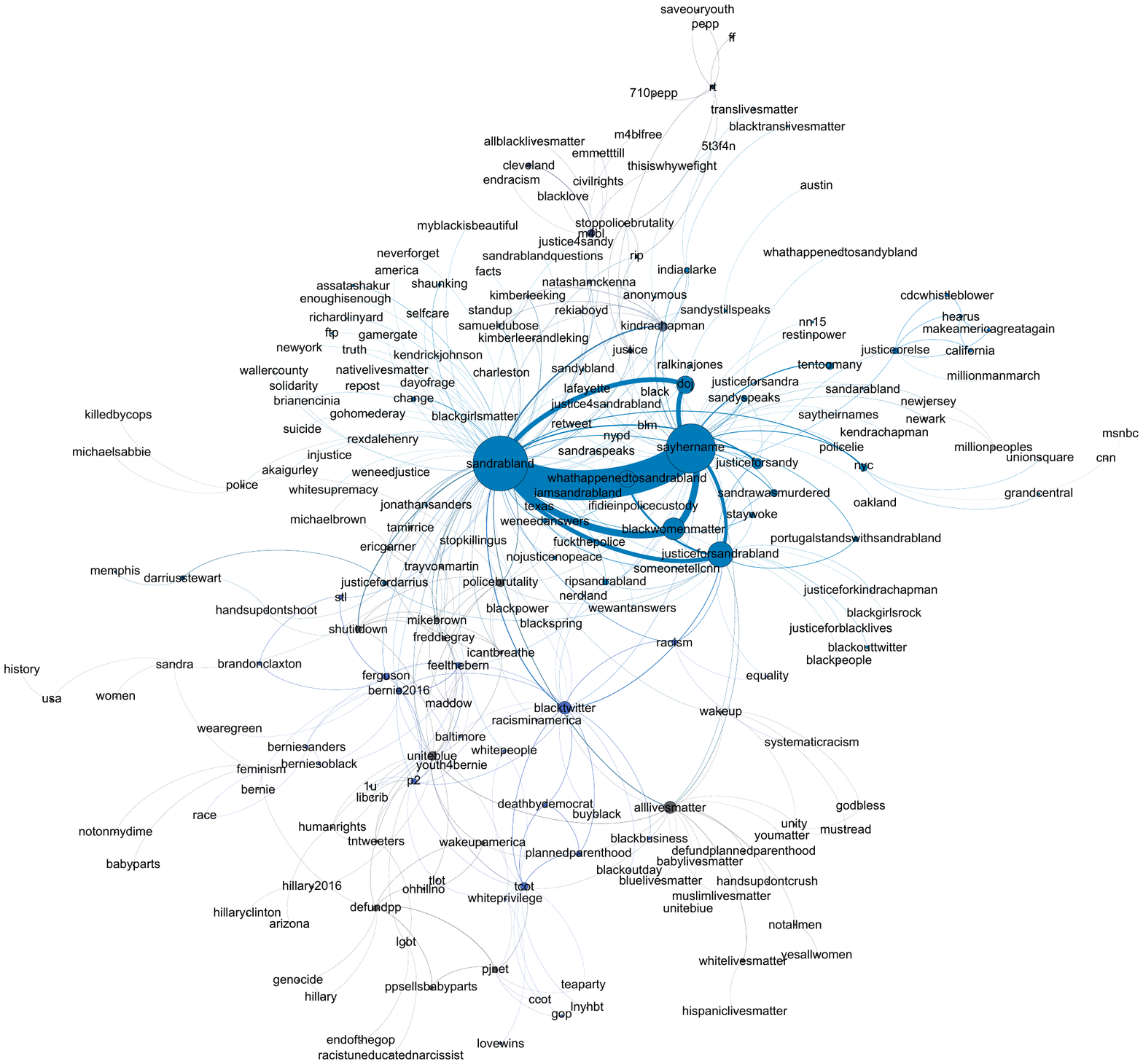}
\caption{\#BlackLivesMatter topic network for the week encapsulating outrage over the death of Sandra Bland.}
\end{figure*}

\begin{figure*}
\includegraphics[scale=.38]{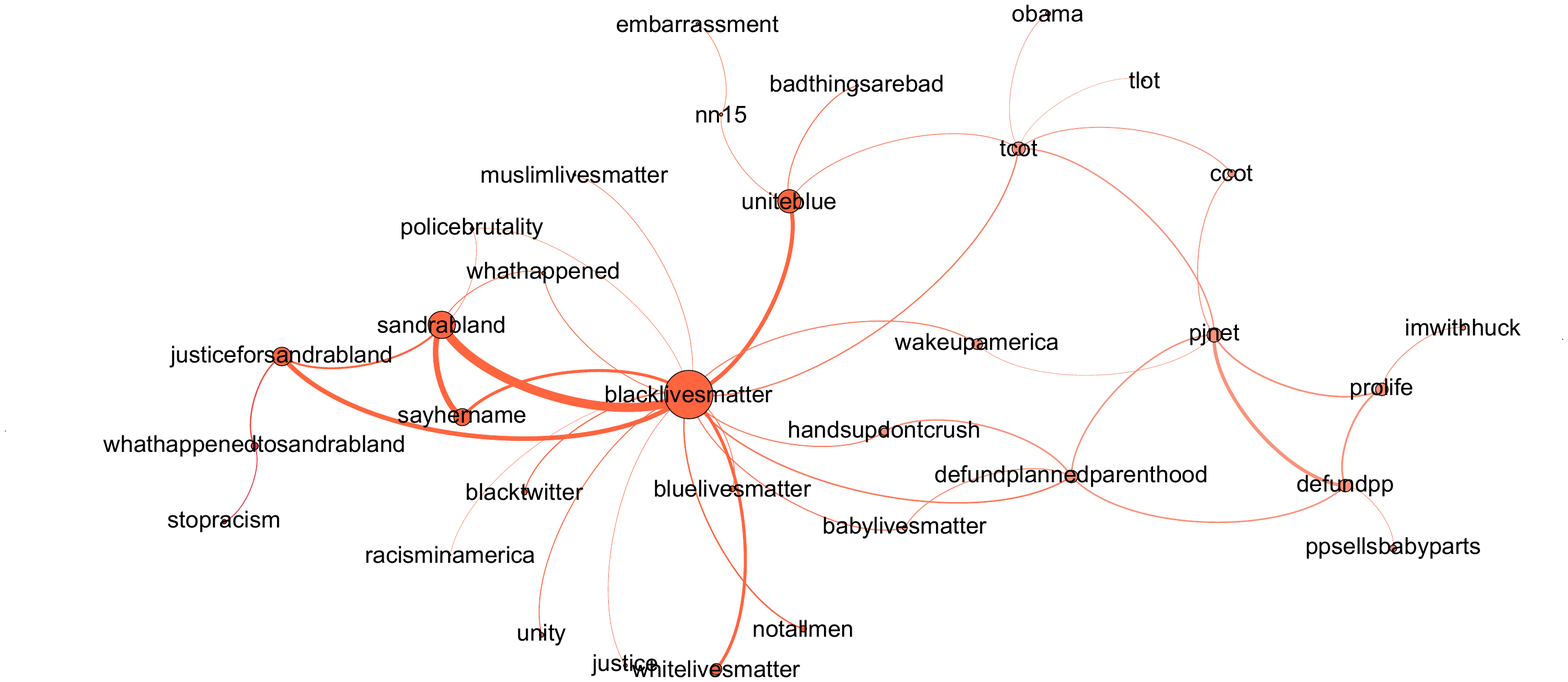}
\caption{\#AllLivesMatter topic network for the week encapsulating outrage over the death of Sandra Bland.}
\label{2015-07-21-backbone-alllives}
\end{figure*}

\end{document}